\documentclass[runningheads]{llncs}

\usepackage{eccv}

\usepackage{eccvabbrv}

\usepackage{graphicx}
\usepackage{booktabs}

\usepackage[accsupp]{axessibility}  

\usepackage{hyperref}

\usepackage{orcidlink}

\usepackage[utf8]{inputenc} 
\usepackage[T1]{fontenc}    
\usepackage{url}            
\usepackage{booktabs}       
\usepackage{amsfonts}       
\usepackage{nicefrac}       
\usepackage{microtype}      
\usepackage{xcolor}         
\usepackage[table]{xcolor}
\usepackage{makecell}
\usepackage{wrapfig}
\usepackage{url}

\usepackage[rightcaption]{sidecap}
\usepackage{graphicx}
\usepackage{subcaption}
\usepackage{booktabs}
\usepackage{amsmath}
\usepackage{amssymb}
\usepackage{float}
\usepackage{enumitem}
\usepackage{multirow}
\usepackage[mathscr]{eucal}
\usepackage[export]{adjustbox}
\usepackage{multicol,lipsum}
\usepackage{arydshln}
\usepackage{pifont}
\usepackage{booktabs}
\usepackage{multirow}
\usepackage{colortbl}
\usepackage{graphicx}
\usepackage{float}

\usepackage[ruled,vlined]{algorithm2e}
\usepackage{pythonhighlight} 
\usepackage{fontawesome5}
\usepackage{marvosym} 

\usepackage[accsupp]{axessibility}  

\def\I{{\bf I}}

\def\q{{\bf q}}

\def\0{{\bf 0}}
\def\1{{\bf 1}}

\def\ph{\mbox{\boldmath$\phi$\unboldmath}}

\DeclareMathOperator*{\argmax}{argmax}

\definecolor{purple}{rgb}{0.56,0.27,0.68}
\definecolor{red}{RGB}{255, 0, 0}
\definecolor{blue_box}{RGB}{145, 202, 233}
\definecolor{yellow_box}{RGB}{255, 230, 154}
\definecolor{purered}{rgb}{1,0,0}
\definecolor{darkblue}{rgb}{0,0,0.8}
\definecolor{lightblue}{rgb}{127,153,240}
\definecolor{grey}{rgb}{0.6,0.6,0.6}
\definecolor{col1}{RGB}{232, 161, 148}
\definecolor{col11}{RGB}{255, 228, 228}
\definecolor{col2}{RGB}{148, 187, 232}
\definecolor{col33}{RGB}{206, 239, 255}
\definecolor{col3}{RGB}{233, 255, 245}
\definecolor{darkgray}{rgb}{0.35,0.35,0.35}
\definecolor{lightgrey}{rgb}{0.85,0.85,0.85}
\definecolor{lightyellow}{RGB}{255,195,78}
\definecolor{lightlightgrey}{rgb}{0.9,0.9,0.9}
\definecolor{verylightBG}{rgb}{0.9,0.99,0.99}
\definecolor{darkgreen}{rgb}{0., 0.85, 0.5}
\definecolor{tfs_lp}{RGB}{130, 176, 210}
\definecolor{tfs_pt}{RGB}{232, 197, 31}
\definecolor{tfs_ft}{RGB}{142, 207, 201}
\definecolor{tfs_ct}{RGB}{250,95,111}

\definecolor{fig2_random}{HTML}{038355}
\definecolor{fig2_entropy}{HTML}{6b9ac8}
\definecolor{fig2_coreset}{HTML}{f8ac8c}
\definecolor{fig2_badge}{HTML}{accf78}
\definecolor{fig2_pcb}{HTML}{c497b2}
\definecolor{fig2_alfa}{HTML}{3d5e80}
\definecolor{fig2_logo}{HTML}{7a8c7b}
\definecolor{fig3_tfs}{HTML}{fa7f6f}

\definecolor{gtred}{RGB}{204, 0, 0}
\definecolor{predgreen}{RGB}{31, 237, 31}
\definecolor{figGreen}{RGB}{56, 118, 29}

\graphicspath{{./figures/}}   

\lstnewenvironment{pythonic}[1][]{\lstset{style=mypython, frame=none, #1}}{}

\begin{document}

\title{Generating a Paracosm for Training-Free Zero-Shot  Composed Image Retrieval} 

\titlerunning{Paracosm}

\author{Tong Wang\inst{1} \and
Yunhan Zhao\inst{2} \and
Shu Kong\inst{1, 3,\text{\Letter}}}

\authorrunning{T.~Wang et al.}


\institute{University of Macau, Macau 999078, China \and
Uiversity of California at Irvine, Irvine, CA 92697, USA \and
Institute of Collaborative Innovation, University of Macau, Macau 999078, China\\
website and code: \url{https://github.com/leowangtong/Paracosm/} 
}

\maketitle

\begin{abstract}
Composed Image Retrieval (CIR),
a task towards personalizing visual intelligence, 
aims to retrieve a target image from a database based on users' multimodal query, which contains a reference image and a modification text.
The text specifies how to alter the reference image to form a ``mental image'',
based on which CIR should find the target image in the database.
The fundamental  challenge of CIR is that this ``mental image'' is not physically available and is only implicitly defined by the query.
The contemporary literature pursues zero-shot methods and uses a Large Multimodal Model (LMM) to generate a textual description for a given multimodal query, and then employs a Vision-Language Model (VLM) for textual-visual matching to search for the target image.
In contrast, we address CIR from first principles by directly generating the ``mental image'' for more accurate matching.
Particularly, we prompt an LMM to generate a ``mental image'' for a given multimodal query and propose to use this ``mental image'' to search for the target image.
As the ``mental image'' has a synthetic-to-real domain gap with real images,
we also generate a synthetic counterpart for each real image in the database to facilitate matching.
In this sense, our method uses LMM to construct a ``paracosm'', where it matches the multimodal query and database images. Hence, we call this method Paracosm. 
Notably, Paracosm is a training-free zero-shot CIR method.
It significantly outperforms existing zero-shot methods on challenging benchmarks, achieving state-of-the-art performance for zero-shot CIR.
  \keywords{Training-Free Zero-Shot Composed Image Retrieval \and Foundation Models \and Synthetic-to-Real Domain Gap}
\end{abstract}

\begin{figure}[t]
  \centering
  \centering
  \vspace{-3mm}
  \includegraphics[width=\textwidth]{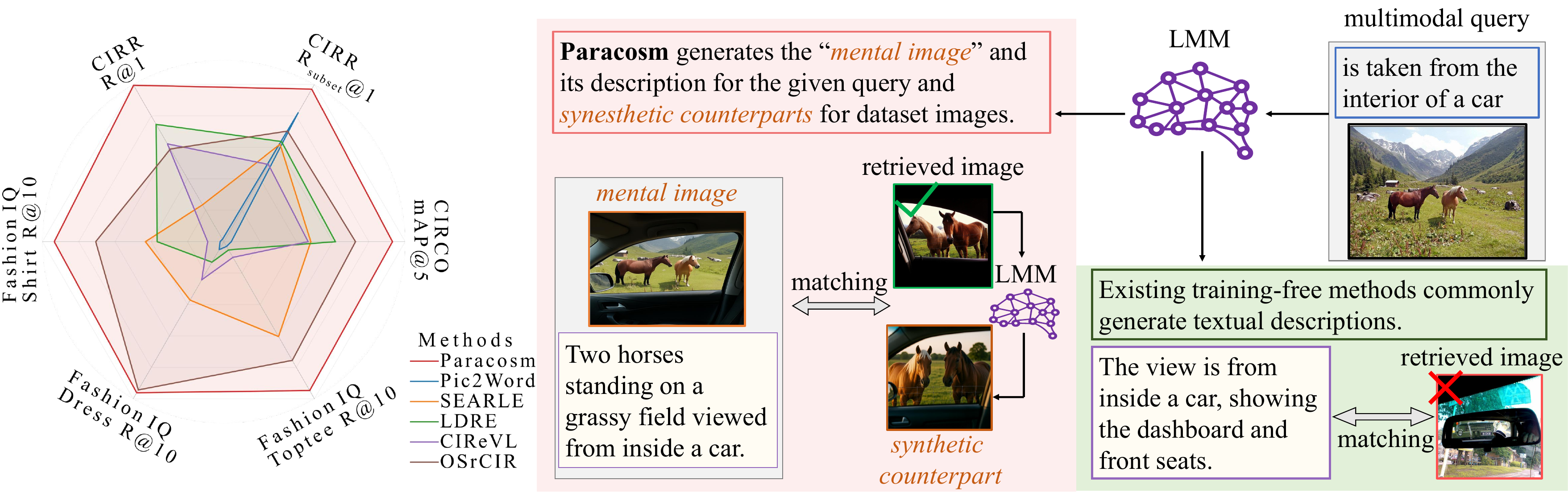}
  \hfill 
  \vspace{-6.5mm}
  \caption{\small {\bf Overview of our method and benchmarking results.}
  Unlike existing training-free methods~\cite{Shyamgopal_2024_CIReVL,yang_2024_ldre,Tang_2025_OSrCIR} generating descriptions for multimodal queries, which use an LMM to generate descriptions for multimodal queries, 
  we use it to generate ``mental images'' for multimodal queries and synthetic counterparts of database images.
  Matching them effectively mitigates synthetic-to-real domain gaps and boosts CIR performance.
  Our final training-free zero-shot method
  \textit{Paracosm} (Fig.~\ref{fig:framework}) significantly outperforms existing zero-shot CIR methods, as summarized in the radar chart on standard benchmarks. 
  Detailed results are provided in \Cref{sec: Detail_results_suppl}.
}
\label{fig:radar-chart}
\vspace{-5mm}
\end{figure}

\section{Introduction}
\label{sec:intro}

Composed Image Retrieval (CIR) formulates new scenarios in search \cite{chen2020image, kwon2024accesslens, arabi2026eurexa} and e-commerce \cite{song2018neural, Wu_2021_FashionIQ},
where it retrieves a target image from a database based on a user-provided multimodal query, which consists of a reference image and modification text \cite{Vo_2019_CIR}. 
The modification text describes how to alter the reference image for which CIR algorithms should find the best-matching image from the database \cite{Liu_2021_CIRR, Baldrati_2023_CIRCO, Wu_2021_FashionIQ}.
It exemplifies an effort for personalized visual intelligence.

{\bf Status Quo.}
CIR was addressed through supervised learning \cite{Baldrati_2022_Combiner, chen2020learning, delmas2022artemis, Liu_2024_BLIP4CIR, Lee_2021_CoSMo} over annotated triplets \textit{<reference image, modification text, target image>}.
As curating large-scale triplet data is prohibitively expensive,
recent works have developed zero-shot CIR (ZS-CIR) approaches \cite{Baldrati_2023_searle, Saito_2023_pic2word, Tang_2025_OSrCIR, yang_2024_ldre}.
Notably, 
``zero-shot'' does not necessarily mean non-learned methods \cite{parashar2024neglected, shen2023high, zhao2024instance, shen2025solving} but emphasizes not directly training models on target dataset.
Rather, they can train models in an indirect way \cite{tang2024context_i2w, Gu_2024_LinCIR, Wang_2025_CIG,zhang2024magiclens, tang2025PrediCIR, huynh2025collm}.
For example, 
many methods \cite{Saito_2023_pic2word, Baldrati_2023_searle} train textual inversion networks on image-text pairs to map images to text tokens for cross-modality matching;
some \cite{Wang_2025_CIG, Li_2025_IPCIR} synthesize a pseudo image for the multimodal query using pretrained image generative models \cite{Rombach_2022_ldm, sauer2024adversarial, Zhou_2024_MIGC, huynh2025collm} to assist with cross-modality matching.
To contrast training-based ZS-CIR approaches,
training-free methods \cite{Shyamgopal_2024_CIReVL, yang_2024_ldre, Tang_2025_OSrCIR} propose to exploit Large Multimodal Models (LMMs) \cite{openai2024gpt4technicalreport, touvron2023llama2, brown2020language} to generate a description for the multimodal query and use it for text-to-image (T2I) matching for CIR.

{\bf Insights.}
The fundamental challenge of CIR is that the multimodal query only implicitly defines a ``mental image'' that does not physically exist to be used 
for retrieving the corresponding target image.
We aspire to address this challenge from first principles by generating a ``mental image'' for a given query to enable more accurate retrieval (\cref{fig:radar-chart}).
While existing works have exploited LMMs to generate textual descriptions,
we leverage LMMs for image generation based on multimodal queries~\cite{wu2025qwenimagetechnicalreport}.
Yet, as the generated ``mental image'' has synthetic-to-real domain gaps compared to the real images in the dataset,
we propose generating a synthetic counterpart for each dataset image, 
which is used to facilitate matching.
As our method essentially uses LMMs to define a synthetic or virtual space, like a  ``paracosm'', we name our method \textbf{Paracosm}.
Notably, \emph{Paracosm is a training-free zero-shot CIR method}.
To validate Paracosm, 
we follow the established experimental protocols that use pretrained Vision-Language Models (VLMs) \cite{radford2021clip, ilharco_gabriel_2021_openclip} for matching.
Extensive experiments on challenging benchmarks demonstrate that Paracosm significantly outperforms existing ZS-CIR methods (ref. a summary in \cref{fig:radar-chart}).

{\bf Contributions.}
We make three major contributions: 
\begin{itemize}
    \item 
    We solve CIR from first principles with the training-free method Paracosm. It generates ``mental images'' for multimodal queries to facilitate matching with database images.
    \item
    We mitigate the synthetic-to-real domain gaps of mental images by generating synthetic counterparts of database images. Matching them together boosts CIR performance.
    \item
    On standard benchmarks, we demonstrate that Paracosm significantly outperforms existing ZS-CIR approaches, achieving state-of-the-art performance.
\end{itemize}

\section{Related Work}
\label{sec:related-work}

{\bf Composed Image Retrieval} (CIR) extends traditional retrieval tasks \cite{shafique2023creating, shafique2024crisp, zhou2025dare}, such as text-to-image retrieval and image-to-image retrieval, by allowing users to use multimodal queries in retrieval~\cite{Vo_2019_CIR}.
CIR was initially approached by supervised learning methods~\cite{Baldrati_2022_Combiner, chen2020learning, delmas2022artemis,Liu_2024_BLIP4CIR,Lee_2021_CoSMo}, i.e., training models over annotated data triplets \textit{<reference image, modification text, target image>}.
These methods \cite{Baldrati_2022_Combiner, zhao2022pl4cir, zhang2024collaborative, xu2023multi, hu2023provla, tian2023fashion} propose to 
train a Transformer or an MLP atop pretrained backbones over the annotated data triplets, intending to fuse the reference image and the modification text in a feature space and to allow matching with database images.
Some methods \cite{levy2024data, Liu_2024_BLIP4CIR,liu2024candidate} finetune a pretrained Vision-Language Model (VLM)~\cite{radford2021clip, li2022blip} for better matching between multimodal queries and database images.
However, as supervised learning methods require costly curation of triplet data, recent methods explore zero-shot CIR (ZS-CIR) \cite{Baldrati_2023_searle, Shyamgopal_2024_CIReVL, Gu_2024_LinCIR, Li_2025_IPCIR, Saito_2023_pic2word, Tang_2025_OSrCIR, Wang_2025_CIG, yang_2024_ldre}. 
We explore \emph{training-free} ZS-CIR and introduce a rather simple method that rivals some recent supervised methods.

{\bf Zero-Shot Composed Image Retrieval} (ZS-CIR) aims to solve CIR without directly training on annotated triplet data \cite{Saito_2023_pic2word}.
It does not mean developing training-free methods but allows training in an indirect manner~\cite{Saito_2023_pic2word, Baldrati_2023_searle, tang2024context_i2w, Gu_2024_LinCIR, zhang2024magiclens,tang2025PrediCIR}.
For example, many methods train a textual inversion network on existing image datasets~\cite{sharma2018cc3m, ILSVRC15imagenet}, mapping the reference image to a pseudo-word token. This token, combined with the modification text, is then used in cross-modality matching with database images.
A couple of recent methods \cite{Wang_2025_CIG, Li_2025_IPCIR} use pretrained generative models~\cite{Rombach_2022_ldm, sauer2024adversarial, Zhou_2024_MIGC} to synthesize a synthetic image (similar to our ``mental image''), termed pseudo-target images, for a given multimodal query. They use these images to augment the textual feature (i.e., output by the textual inversion network) to compute similarity scores with database images.
Importantly, training-free ZS-CIR methods \cite{Shyamgopal_2024_CIReVL, yang_2024_ldre, Tang_2025_OSrCIR, Sun_2025_cotmr, cheng2025autocir}
have emerged that leverage an LMM to generate a description for a given multimodal query. This avoids training a separate textual inversion network.
Notably, existing ZS-CIR methods have not considered generating either descriptions or synthetic images for database images.
In contrast, our work does so, motivated to address ZS-CIR from first principles. Specifically, we propose to generate the ``mental image'' for the query and synthetic counterparts of real database images, mitigating synthetic-to-real domain gaps for better performance.


{\bf Foundation Models} (FMs), pretrained on various formats of web-scale data across multiple modalities, demonstrate unprecedented performance on downstream tasks in a zero-shot manner.
Consequently, 
FMs are extensively utilized in the CIR literature~\cite{li2024improving, bao2025MLLM-I2W, yang2024sda, liu2024zeroshotcomposedtextimageretrieval, Brooks_2023_InstructPix2Pix}. 
First, Vision-Language Models (VLMs), pretrained on web-scale image-text pairs \cite{radford2021clip, ilharco_gabriel_2021_openclip}, are commonly employed in CIR to extract features from modification texts and images, enabling cross-modality matching \cite{parashar2024neglected,
sun2024alpha, Shyamgopal_2024_CIReVL, Tang_2025_OSrCIR, Baldrati_2022_Combiner, Li_2025_IPCIR}.
Second, Large Language Models (LLMs), pretrained on massive text corpora \cite{tan2019lxmert, brown2020language, touvron2023llama, bai2023qwen},
are utilized in CIR to generate a target image description by incorporating the reference image caption and the modification text \cite{yang_2024_ldre, Li_2025_IPCIR}.
Third, Large Multimodal Models (LMMs)~\cite{openai2024gpt4technicalreport, wu2025qwenimagetechnicalreport, team2023gemini, bai2025qwen2}, which are pretrained on web-scale multimodal data and typically larger than VLMs in parameter size, 
are used in CIR to generate image captions \cite{Shyamgopal_2024_CIReVL, yang_2024_ldre, Li_2025_IPCIR} for multimodal queries.
Earlier CIR works leverage LMMs to create triplet data for supervised training \cite{liu2024zeroshotcomposedtextimageretrieval, zhang2024magiclens, tang2025PrediCIR}. 
In contrast, most recent training-free ZS-CIR methods \cite{Shyamgopal_2024_CIReVL, Tang_2025_OSrCIR, yang_2024_ldre} employ LMMs to generate textual descriptions for multimodal queries and utilize a VLM to match these descriptions with database images, transforming the CIR task into a text-to-image retrieval problem.
Building upon this foundation, we extensively exploit LMMs not only to generate ``mental images'' for multimodal queries but also to create synthetic counterparts of real database images.
Matching them effectively mitigates synthetic-to-real domain gaps, thereby significantly enhancing CIR performance.

\section{Methodology}
\label{sec:method}
\begin{figure*}[t]
  \centering
  \includegraphics[width=1\linewidth, page=1, clip=true,trim = 0mm 0mm 0mm 0mm]{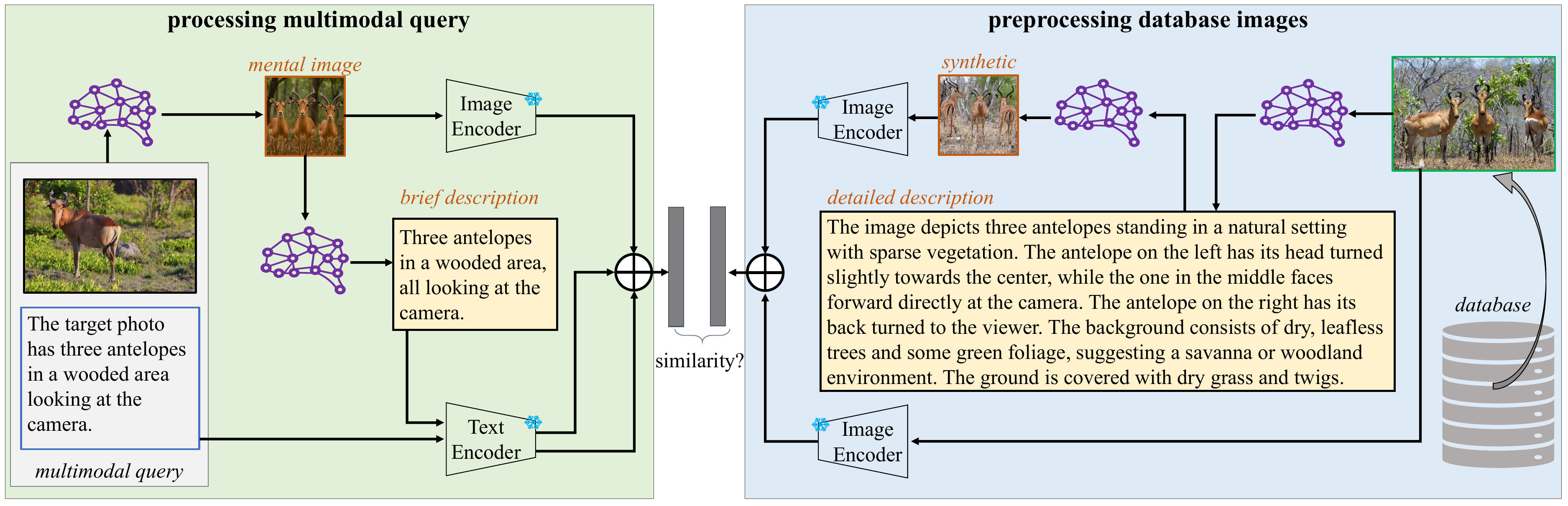}
  \hfill
  \vspace{-6.5mm}
  \caption{\small
  {\bf Flowchart of our training-free zero-shot CIR method Paracosm.} 
  Given a multimodal query that consists of a reference image and a modification text, we feed it to an LMM to generate a ``mental image''.
  We further generate a brief description for it.
  Both the ``mental image'' and description, as well as the modification text, are used as feature representations for the query. 
  As the ``mental image'' is synthetic, we mitigate synthetic-to-real domain gaps by generating synthetic counterparts of database images. To do so, we use the LMM to generate \emph{detailed} descriptions, which are used as prompts for image generation.
  For a database image,
  we use both itself (i.e., the real photo) and its synthetic counterpart as representation for retrieval.
  In sum, our method uses LMMs to create a virtual {\bf paracosm}, where it matches the query and database images.
} 
\label{fig:framework}
\vspace{-3mm}
\end{figure*}

We begin by defining the ZS-CIR problem and its development protocol. 
We then introduce our {\em training-free} method, Paracosm. 
Despite its conceptual simplicity, Paracosm achieves state-of-the-art performance among zero-shot methods.
Fig.~\ref{fig:framework} illustrates the complete pipeline.

\subsection{Preliminaries of Zero-Shot CIR }
\label{ssec:prelim}
{\bf Task Definition.}
Let $(\I_{ref}, {\mathbf t}_{mod})$ represent the reference image and modification text of a given multimodal query.
${\mathbf t}_{mod}$ describes how to alter $\I_{ref}$ to match what the user wants to search for, i.e., a target image $\I_{target}$ from a given database consisting of $n$ images $\{\I^1, \I^2, ..., \I^n\}$.
ZS-CIR aims to develop methods to retrieve target images for multimodal queries without directly training on annotated triplets $(\I_{ref}, {\mathbf t}_{mod}, \I_{target})$.

{\bf Methodology Development.}
While ZS-CIR eschews a training set of annotated triplet data, it still allows exploiting data available in the open world and pretrained FMs therein.
Following this protocol, existing ZS-CIR methods leverage LMMs in different ways, e.g., using a VLM for cross-modality matching~\cite{Saito_2023_pic2word, Baldrati_2023_searle, Gu_2024_LinCIR} and an LMM for generating descriptions \cite{Shyamgopal_2024_CIReVL, Tang_2025_OSrCIR, yang_2024_ldre}.
By exploiting open-world data \cite{ILSVRC15imagenet, sharma2018cc3m},
some methods train necessary models for fusing $\I_{ref}$ and ${\mathbf t}_{mod}$ into features \cite{Saito_2023_pic2word, Baldrati_2023_searle, Gu_2024_LinCIR},
or for generating pseudo images for the multimodal query \cite{Wang_2025_CIG, Li_2025_IPCIR}.
In this work, we aspire to develop a \emph{training-free} method by extensively leveraging LMMs, without training any new models.
Next, we present our method Paracosm in detail.

\subsection{The Proposed Method: Paracosm}
\label{ssec:paracosm}
Paracosm is a rather simple method that processes multimodal queries and database images using LMMs.
Below, we describe how it processes them and matches them for retrieval.

\begin{figure}[t]
\centering
\includegraphics[width=1\linewidth, clip=true,trim = 0mm 0mm 0mm 0mm]{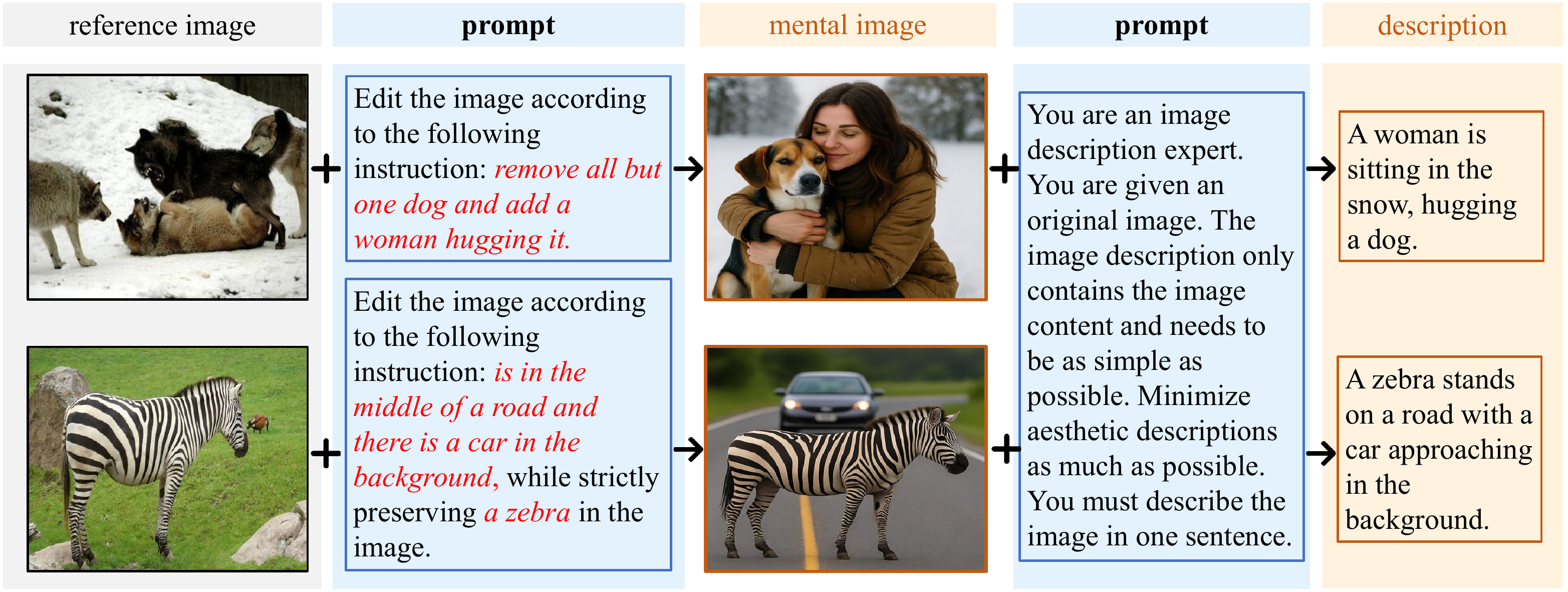}
\vspace{-6.5mm}
\caption{\small
  {\bf Illustration of processing a multimodal query.}
  Two random examples from CIRR and CIRCO datasets are displayed in the two rows, respectively.
  For a multimodal query consisting of a reference image and  \textcolor{red}{\textit{modification texts}},
  we employ a prompt incorporating the latter to edit the former, generating a mental image representing this  query.
  To adapt to the varying modification text across benchmarks, where a modification text can contain a relative caption and a shared concept (ref. CIRCO in the second row),
  we design the prompt to incorporate both.
  Further, for the mental image,
  we prompt an LMM to generate a a concise, single-sentence description, exclusively focusing on its visual content while minimizing aesthetic details.
  We use both mental image and short description to retrieve the target image from the database.
  }
\label{fig:prompt}
\vspace{-4.5mm}
\end{figure}

\begin{figure*}[t]
\centering
\includegraphics[width=1\linewidth, clip=true,trim = 0mm 0mm 0mm 0mm]{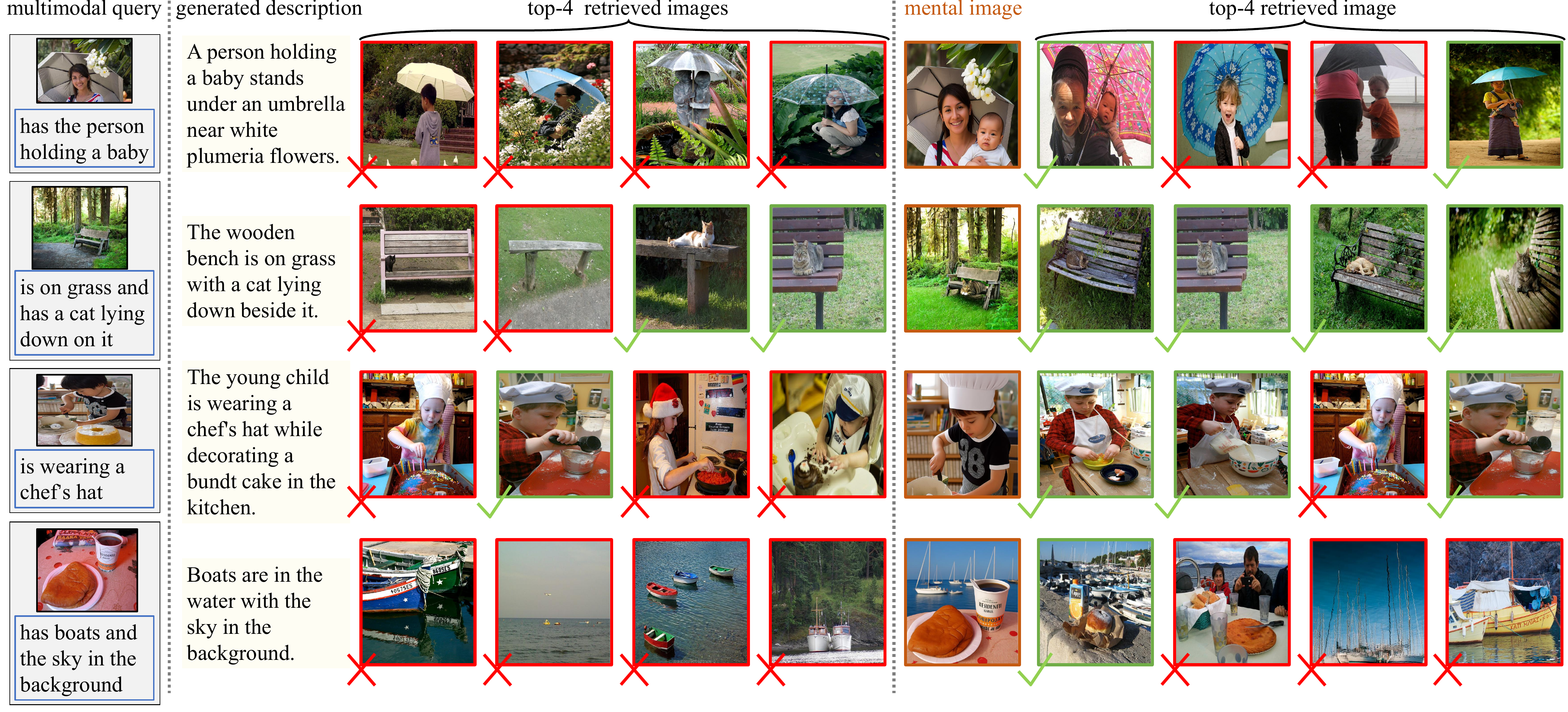}
\vspace{-6.5mm}
  \caption{\small 
  {\bf Comparison of qualitative results} between OSrCIR~\cite{Tang_2025_OSrCIR} and our Paracosm. 
  We show four examples from the CIRCO dataset~\cite{Baldrati_2023_CIRCO} in the first column, followed by generated descriptions and top-4 retrievals by OSrCIR, and the mental images and top-4 retrievals by Paracosm.
  For each multimodal query, OSrCIR uses an LMM to generate a description, uses it to match database images, and returns top-ranked ones.
  Instead, Paracosm uses an LMM to generate a ``mental image'' for each query, which contains much richer information than a description, allowing image-to-image matching for better retrieval.  
  Consequently, Paracosm yields better retrievals than OSrCIR.
}
\label{fig:fig_generated_img_has_more_visual_informantion}
\vspace{-5mm}
\end{figure*}

{\bf Processing Multimodal Query.}
For a multimodal query consisting of a reference image and modification text, 
Paracosm first generates the ``mental image'' $\I_{mental}$ using an LMM~\cite{wu2025qwenimagetechnicalreport}.
Based on this mental image $\I_{mental}$, we further generate a brief textual description ${\mathbf t}_{query}$, 
which can be thought of as the description of the target image $\I_{target}$.
For this step, we use a prompt that instructs the LMM \cite{qwen2.5-VL} to focus solely on visual content while minimizing aesthetic details, producing single-sentence descriptions. 
Fig.~\ref{fig:prompt} illustrates the prompt templates used for processing a multimodal query.
 Since different datasets provide modification texts in varying 
formats, we adapt the prompt template structure accordingly to ensure proper input formatting for the LMM.
Given a multimodal query,
Paracosm uses both the generated mental image and short description for retrieval.
This is different from some recent methods \cite{Shyamgopal_2024_CIReVL, Tang_2025_OSrCIR, yang_2024_ldre},
which exclusively use generated descriptions for retrieval.
For example, 
some of such methods first generate a description for the reference image and then revise the description with the modification text using an LLM.
\cref{fig:fig_generated_img_has_more_visual_informantion} demonstrates that solely relying on generated descriptions misses crucial visual information, 
whereas our mental images retain rich information, facilitating CIR.

\begin{figure}[t]
\centering
\includegraphics[width=1\linewidth, clip=true,trim = 0mm 0mm 0mm 0mm]{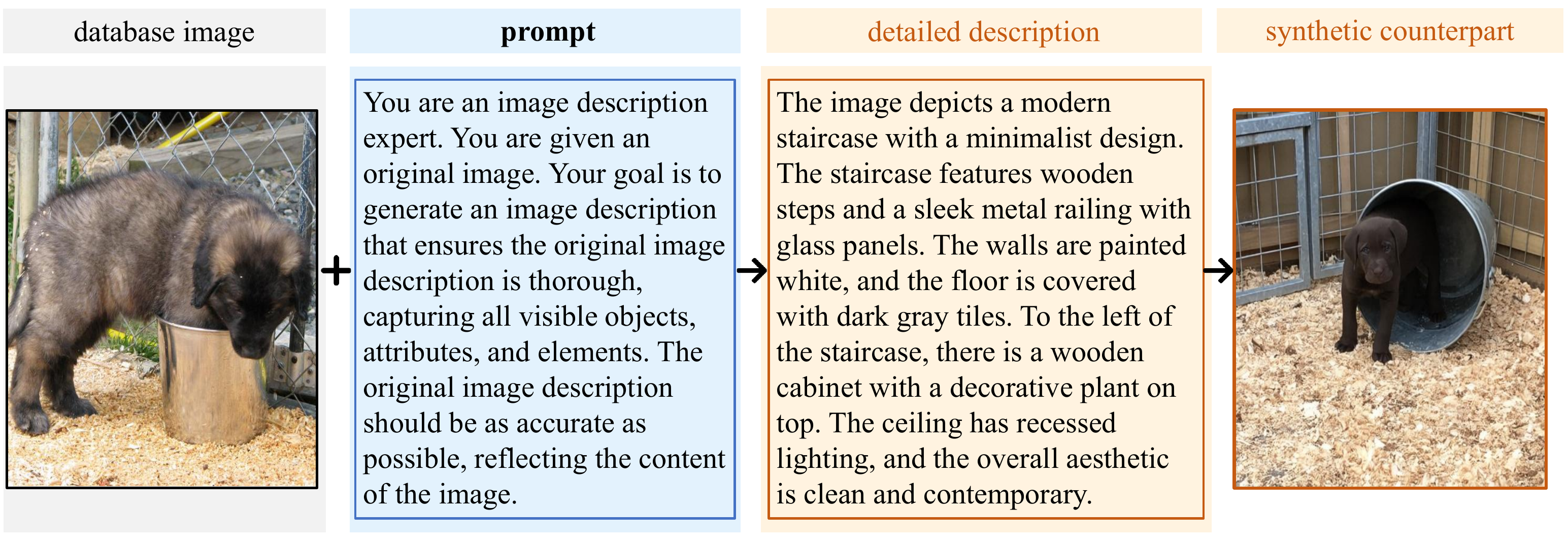}
\vspace{-6.5mm}
\caption{\small
  {\bf Illustration of processing database images.}  
  For a database image,
  we first prompt an LMM to generate a comprehensive description about its visual content,
  capturing all visible objects, attributes, and visual elements.
  Using this description,  
  we prompt a text-to-image generation model
  to produce a synthetic counterpart for this database image.
  For a database image,
  we use both itself (i.e., the real photo) and its synthetic counterpart as representation for retrieval.
  }
\label{fig:synthetic_counterpart_prompt}
\vspace{-5mm}
\end{figure}

{\bf Pre-Processing Database Images.}
As the generated mental images are synthetic and have synthetic-to-real domain gaps, directly matching them with real images from the database is suboptimal.
To mitigate this gap, we create synthetic counterparts for database images and use such counterparts to assist matching~\cite{zhao2020domain, gao2023back, guo2025everything}.
Specifically, 
Paracosm first leverages an LMM~\cite{qwen2.5-VL} to generate a \emph{detailed} description for each database image.
We employ a description prompt that emphasizes capturing 
all visible objects, attributes, spatial relationships, and fine-grained visual elements to ensure maximum fidelity.
Fig.~\ref{fig:synthetic_counterpart_prompt} illustrates this prompt template. 
The detailed description then serves as a prompt for a text-to-image (T2I) generation model~\cite{wu2025qwenimagetechnicalreport}, generating a synthetic counterpart $\I_{syn}^i$ for each database image $\I^{i}$.
Both real database images and synthetic counterparts are used jointly for matching multimodal queries.

{\bf Matching for Retrieval.}
After transforming both multimodal queries and database images into the virtual paracosm, we construct features using a pretrained VLM, which consists of a visual encoder $V(\cdot)$ and a text encoder $T(\cdot)$.
Specifically, the query feature $\q$ and the $i^{th}$ database image feature $\ph^i$ are computed as follows:
\begin{equation}\small
\begin{split}
& {\mathbf q} = \lambda (V(\I_{mental}) + T({\mathbf t}_{query})) + (1 - \lambda){T({\mathbf t}_{mod})} 
\\
& \ph^i = V(\I^i) + V(\I_{syn}^i) 
\end{split}
\label{eq:features}
\end{equation}
where $\lambda$ is a hyperparameter controlling the contribution of incorporating the modification text ${\mathbf t}_{mod}$.
The weights of features are discussed in detail in \Cref{ssec:ablation} and \Cref{sec:further_analysis_suppl}. 
Finally, Paracosm computes the cosine similarity score and returns the index of the potential target image:
\begin{equation}\small
    i^{*} = \argmax_{i=1 \cdots n} 
            \frac{{\mathbf q}^T \ph^i}{ \|{\mathbf q}\|_2   \|\ph^i\|_2 }
\nonumber
\end{equation}

\subsection{Remarks}
\label{ssec:remarks}

{\bf Paracosm makes better use of LMMs.} 
Existing ZS-CIR approaches~\cite{yang_2024_ldre, Tang_2025_OSrCIR, Shyamgopal_2024_CIReVL} also leverage LMMs but focus on generating descriptions for multimodal queries and transforming the CIR problem into a text-to-image retrieval problem.
However, text descriptions alone cannot sufficiently capture the rich visual information crucial to CIR and hence often lead to incorrectly retrieved images, as demonstrated in \cref{fig:fig_generated_img_has_more_visual_informantion}. 
A couple of recent works~\cite{Li_2025_IPCIR, Wang_2025_CIG} realize the importance of incorporating pseudo-target images to facilitate CIR, e.g., by exploiting a pretrained text-to-image (T2I) generative model~\cite{Zhou_2024_MIGC} to synthesize a pseudo-target image based on a description.
Differently, we leverage an LMM that has image editing capability, i.e., editing the reference image based on the modification text into a ``mental image''. 
As shown in \Cref{tab:further_analysis}, editing reference images yields better CIR results than T2I generation of pseudo-target images.
Moreover, Paracosm exploits an LMM to generate detailed descriptions for each database image and uses these descriptions as prompts to generate synthetic counterparts.

{\bf Is CIR still needed given the high-quality edited images?} 
This question emerges in front of the high quality of the generated ``mental images'' based on multimodal queries (\cref{fig:fig_generated_img_has_more_visual_informantion}).
We argue that CIR is still important in real-world visual search applications.
For example,
in e-commerce and the fashion industry,
users might want to start from a photo of clothes and search for a different genre or style in an e-shop by specifying how to alter the photo.
That said, no matter how photorealistic a mental image is, image generation cannot replace a real product in inventory.

{\bf Computation cost.}
As Paracosm extensively exploits LMMs, it has a high computation cost, 
especially for generating images, i.e., the mental images for the query and synthetic counterparts of database images.
Yet, this is not a flaw pertaining only to Paracosm,
as existing works also rely on generative models for retrieval. 
For example, 
prevailing ZS-CIR methods use LMMs to generate descriptions \cite{yang_2024_ldre, Tang_2025_OSrCIR, Shyamgopal_2024_CIReVL, Li_2025_IPCIR},
and some turn to generative models to synthesize images \cite{Li_2025_IPCIR, Wang_2025_CIG}.
Importantly, like these methods, 
Paracosm can process database images ahead of time and does not require computing features for them on the fly during inference.
Thus, the major computational overhead is on the process of a given multimodal query during inference, including generating the corresponding mental images and their descriptions.
As efficient inference in generative models is an important topic and has been greatly advanced through model optimization~\cite{yin2024one, baldridge2024imagen} and optimized implementation~\cite{diffsynth-engine2025, flux-2-2025}, Paracosm would not suffer from computation in the long run.

{
\begin{table}[t]
\centering
\small
\caption{\small 
\textbf{Comparing methods for mental image generation.} 
We adopt the CLIP ViT-B/32 VLM and report results on the CIRR test set. 
``T2I Generation'' means that we first generate pseudo-target descriptions and then use them to generate mental images, as done in \cite{Li_2025_IPCIR}.
``Image Edit'' is our final design choice,
for generating mental images in Paracosm,
which directly edits the  reference image based on the corresponding modification text.
Clearly, the latter performs better.
Moreover, for ``Image Edit'', we compare using Qwen-Image-Edit vs. LongCat-Image-Edit as the image generator.
Results show stable performance of our Paracosm.
}
\vspace{-3mm}
\setlength{\tabcolsep}{2.0em}
\scalebox{0.9}
{
    \begin{tabular}{lccccccccccccc}
    \toprule

    \textbf{Method} & R@1 &  R@5  & R@10 &  R@50 \\
    \cmidrule(lr){1-1}
    \cmidrule(lr){2-5}
    
    T2I Generation  & 31.71 & 61.37 & 73.59 & 91.54 \\

    Image Edit w/ Qwen & \textbf{32.27} & \textbf{62.60} & \textbf{75.16} & \textbf{92.60} \\

    Image Edit w/ LongCat & 32.12 & 62.20 & 74.43 & 92.24 \\
    
\bottomrule
\end{tabular}
}
\vspace{-5mm}
\label{tab:further_analysis}
\end{table}
}

{
\begin{table*}[t]
\centering
\small
\caption{\small
\textbf{Benchmarking results on CIRCO and CIRR test sets.}
We evaluate Paracosm against state-of-the-art zero-shot CIR methods 
using ViT-L/14 and ViT-G/14 backbones.
Metrics include Recall@k/Recall$_{\text{Subset}}$@k for CIRR and mAP@k for CIRCO (which contains multiple target images per query). 
Best results are in \textbf{bold}, second-best are \underline{underlined}.
Paracosm achieves state-of-the-art performance among zero-shot methods across all 
benchmarks and backbones, demonstrating the effectiveness of our generation approach.
Notably, our training-free method even surpasses several 
supervised approaches, highlighting its practical potential.
}
\vspace{-3mm}
\setlength{\tabcolsep}{0.4em}
\scalebox{0.67}
{
    \begin{tabular}{lllccccccccccccc}
    \toprule
    \multicolumn{3}{c}{} 
    & \multicolumn{6}{c}{CIRR} & \multicolumn{4}{c}{CIRCO}\\
    \cmidrule(lr){4-9}
    \cmidrule(lr){10-13}
    & &
    & \multicolumn{3}{c}{Recall@k} 
    & \multicolumn{3}{c}{Recall$_{\text{Subset}}$@k} 
    & \multicolumn{4}{c}{mAP@k}  

    \\
   
     \textbf{Backbone} & \textbf{Method} &  \textcolor{gray}{venue\&year} & k=1 &  k=5  & k=10  & k=1 &  k=2  & k=3 &  k=5 & k=10 &  k=25 & k=50 \\
    \cmidrule(lr){1-1}
    \cmidrule(lr){2-2}
    \cmidrule(lr){3-3} 
    \cmidrule(lr){4-6}
    \cmidrule(lr){7-9}
    \cmidrule(lr){10-13} 

    \multirow{2}{*}{supervised} & Combiner~\cite{Baldrati_2022_Combiner} & \textcolor{gray}{CVPR'22} & 33.59 & 65.35 & 77.35 & 62.39 & 81.81 & 92.02 & -- & -- & -- & -- \\
    
    \multirow{2}{*}{methods} & BLIP4CIR~\cite{Liu_2024_BLIP4CIR}  & \textcolor{gray}{WACV'24} & 40.17 & 71.81 & 83.18 & 72.34 & 88.70 & 95.23 & -- & -- & -- & -- \\
    
    & CLIP-ProbCR~\cite{li2024clipProbCR}  & \textcolor{gray}{ICMR'24} & 23.32 & 54.36 & 68.64 & 54.32 & 76.30 & 88.88 & -- & -- & -- & -- \\

    \midrule 
    
    \multirow{11}{*}{ViT-L/14} & Image-only  & \textcolor{gray}{baseline} & \ \ 7.47 & 23.86 & 34.10 & 20.82 & 41.88 & 61.23 & \ \ 1.80 & \ \ 2.43 & \ \ 3.04 & \ \ 3.45 \\

    & Text-only  & \textcolor{gray}{baseline} & 22.05 & 45.64 & 57.54 & 61.69 & 80.31 & 90.41 & \ \ 2.99 & \ \ 3.18 & \ \ 3.68 & \ \ 3.92 \\

    & Image+Text  & \textcolor{gray}{baseline} & 10.58 & 32.65 & 45.69 & 31.08 & 55.71 & 73.90 & \ \ 3.89 & \ \ 4.79 & \ \ 5.93 & \ \ 6.47 \\

    & Pic2Word~\cite{Saito_2023_pic2word}  & \textcolor{gray}{CVPR'23} & 23.90 & 51.70 & 65.30 & 53.76 & 74.46 & 87.07 & \ \ 8.72 & \ \ 9.51 & 10.64 & 11.29 \\
									
    & SEARLE~\cite{Baldrati_2023_searle}  & \textcolor{gray}{ICCV'23} & 24.24 & 52.48 & 66.29 & 53.76 & 75.01 & 88.19 & 11.68 & 12.73 & 14.33 & 15.12 \\

    & LinCIR~\cite{Gu_2024_LinCIR}  & \textcolor{gray}{CVPR'24} & 25.04 & 53.25 & 66.68 & 57.11 & 77.37 & 88.89 & 12.59 & 13.58 & 15.00 & 15.85 \\

    & LDRE~\cite{yang_2024_ldre}  & \textcolor{gray}{SIGIR'24} & 26.53 & 55.57 & 67.54 & 60.43 & 80.31 & 89.90 & 23.35 & 24.03 & 26.44 & 27.50 \\
    
    & CIReVL~\cite{Shyamgopal_2024_CIReVL}  & \textcolor{gray}{ICLR'24} & 24.55 & 52.31 & 64.92 & 59.54 & 79.88 & 89.69 & 18.57 & 19.01 & 20.89 & 21.80 \\

    & IP-CIR + LDRE~\cite{Li_2025_IPCIR} & \textcolor{gray}{CVPR'25} & \underline{29.76} & \underline{58.82} & \underline{71.21} & \underline{62.48} & \underline{81.64} & \underline{90.89} & \underline{26.43} & \underline{27.41} & \underline{29.87} & \underline{31.07} \\

    & CIG + SEARLE~\cite{Wang_2025_CIG}  & \textcolor{gray}{CVPR'25} & 26.72 & 55.52 & 68.10 & 57.95 & 77.81 & 89.45 & 12.84 & 13.64 & 15.32 & 16.17 \\
    
    & \bf Paracosm  & \textcolor{gray}{ours} & \textbf{31.95} & \textbf{61.56} & \textbf{72.96} & \textbf{64.68} & \textbf{82.89} & \textbf{91.47} & \textbf{30.24} & \textbf{31.51} & \textbf{34.29} & \textbf{35.42} \\

    \midrule
    \multirow{9}{*}{ViT-G/14} & Pic2Word~\cite{Saito_2023_pic2word}  & \textcolor{gray}{CVPR'23} & 30.41 & 58.12 & 69.23 & 68.92 & 85.45 & 93.04 & \ \ 5.54 & \ \ 5.59 &  \ \ 6.68 &  \ \ 7.12 \\

    & SEARLE~\cite{Baldrati_2023_searle}  & \textcolor{gray}{ICCV'23} & 34.80 & 64.07 & 75.11 &  68.72 & 84.70 & 93.23 & 13.20 & 13.85 & 15.32 & 16.04 \\

    & LinCIR~\cite{Gu_2024_LinCIR}  & \textcolor{gray}{CVPR'24} & 35.25 & 64.72 & 76.05 & 63.35 & 82.22 & 91.98 & 19.71 & 21.01 & 23.13 & 24.18 \\

    & LDRE~\cite{yang_2024_ldre}  & \textcolor{gray}{SIGIR'24} & 36.15 & 66.39 & 77.25 & 68.82 & 85.66 & 93.76 & 31.12 & 32.24 & 34.95 & 36.03 \\
    
    & CIReVL~\cite{Shyamgopal_2024_CIReVL}  & \textcolor{gray}{ICLR'24} & 34.65 & 64.29 & 75.06 & 67.95 & 84.87 & 93.21 & 26.77 & 27.59 & 29.96 & 31.03 \\

    & CIG + LinCIR~\cite{Wang_2025_CIG}  & \textcolor{gray}{CVPR'25} & 36.05 & 66.31 & 76.96 & 64.94 & 83.18 & 91.93 & 20.64 & 21.90 & 24.04 & 25.20 \\

    & CoTMR~\cite{Sun_2025_cotmr} & \textcolor{gray}{ICCV'25} & 36.36 & \underline{67.52} & \underline{77.82} & \textbf{71.19} & \underline{86.34} & \underline{93.87} & \underline{32.23} & \underline{32.72} & \underline{35.60} & \underline{36.83}\\
    
    & OSrCIR~\cite{Tang_2025_OSrCIR} & \textcolor{gray}{CVPR'25} & \underline{37.26} & 67.25 & 77.33 & 69.22 & 85.28 & 93.55 & 30.47 & 31.14 & 35.03 & 36.59 \\

    & \bf Paracosm  & \textcolor{gray}{ours} & \textbf{39.30} & \textbf{70.41} & \textbf{80.39}  & \underline{70.82} & \textbf{86.92} & \textbf{94.46} & \textbf{39.82} & \textbf{40.86} & \textbf{43.96} & \textbf{45.05} \\
    
\bottomrule

\end{tabular}
}
\vspace{-5mm}
\label{tab:cirr_circo_results} 
\end{table*}
}

\section{Experiments}
\label{sec:experiments}

We conduct extensive experiments to validate the proposed Paracosm, comparing it against existing methods and ablating its components. 
We start by introducing datasets, metrics, implementation details, and compared methods.

{\bf Datasets.}
Following the literature \cite{Saito_2023_pic2word, Baldrati_2023_searle, cheng2025autocir, yang_2024_ldre, Li_2025_IPCIR},
we use established benchmarks:
CIRR~\cite{Liu_2021_CIRR}, CIRCO~\cite{Baldrati_2023_CIRCO}, and Fashion IQ~\cite{Wu_2021_FashionIQ}. 
These datasets are publicly available for non-commercial research and educational purposes.
CIRCO is released under the CC-BY-NC 4.0 license;
CIRR is licensed under CC-BY 4.0;
Fashion-IQ is distributed under the Community Data License Agreement (CDLA). 
As we focus on ZS-CIR, we do not exploit their training data to develop models.
For Fashion IQ, which only releases its validation set, we benchmark methods on this val-set; for other datasets, we benchmark methods on their official test sets. Moreover, Fashion IQ has multiple subsets for testing,
we report results on each of them.
We provide more details in the supplementary \Cref{sec:datasets_suppl}.

{\bf Metrics.}
Following prior arts~\cite{Shyamgopal_2024_CIReVL, Tang_2025_OSrCIR, yang_2024_ldre, Gu_2024_LinCIR},
we use Recall@k (R@k) as the metric for CIRR and Fashion IQ.
For CIRCO, we use mean average precision (mAP@k) as it has more target images for queries.

{\bf Implementation Details.}
Following the literature~\cite{Shyamgopal_2024_CIReVL, Tang_2025_OSrCIR, yang_2024_ldre}, 
we report the results of all the methods using CLIP ViT-L/14 \cite{radford2021clip} and OpenCLIP ViT-G/14 \cite{ilharco_gabriel_2021_openclip} to compute image and text  features.
The LMMs used in Paracosm are Qwen2.5-VL-7B-Instruct~\cite{qwen2.5-VL}, Qwen-Image~\cite{wu2025qwenimagetechnicalreport}, and Qwen-Image-Edit~\cite{wu2025qwenimagetechnicalreport}, for description generation, database image synthesis, and ``mental image'' generation.
For all image generation tasks (mental images and synthetic counterparts), we set the output resolution to $512 \times 512$ 
pixels, with all other parameters following the default values recommended by the official implementations.
We preprocess database images with a computing cluster of 16$\times$ NVIDIA A100 GPUs.
We run individual methods on a single NVIDIA A100 GPU.
Figures \ref{fig:prompt} and \ref{fig:synthetic_counterpart_prompt} show the detailed generation process.
Paracosm 
(\cref{fig:framework}) employs a hyperparameter $\lambda$ (Eq.~\ref{eq:features}) to control the contribution of modification texts from the query. We tuned $\lambda$ on the CIRR validation set using CLIP ViT-B/32 and set $\lambda=0.3$ throughout our work.
\cref{fig:lambda} studies its effects on the final CIR performance over the benchmarks. 
Interestingly, $\lambda=0.3$ consistently produces the highest numeric metrics on all the benchmarks.

\begin{table*}[t]
\centering
\small
\caption{\small
\textbf{Benchmarking results on the Fashion IQ validation set.}
We compare Paracosm against zero-shot CIR methods using ViT-L/14 and ViT-G/14 backbones. 
Metrics include Recall@10 (R@10) and Recall@50 (R@50) across three categories (Shirt, Dress, Toptee) and their average.
Best results are in \textbf{bold}, second-best are \underline{underlined}.
Paracosm achieves state-of-the-art performance among zero-shot methods across all 
benchmarks and backbones.
}
\vspace{-3mm}
\setlength{\tabcolsep}{0.57em}
\scalebox{0.68}
{
    \begin{tabular}{lllcccccccccccc}
    \toprule
    \multicolumn{3}{c}{} 
    & \multicolumn{2}{c}{Shirt} & \multicolumn{2}{c}{Dress} & \multicolumn{2}{c}{Toptee} & \multicolumn{2}{c}{Average} \\
    \cmidrule(lr){4-5}
    \cmidrule(lr){6-7}
    \cmidrule(lr){8-9}
    \cmidrule(lr){10-11}

     \textbf{Backbone} & \textbf{Method} &  \textcolor{gray}{venue\&year} & R@10 &  R@50  & R@10 &  R@50  & R@10 &  R@50  & R@10 &  R@50 \\
    \cmidrule(lr){1-1}
    \cmidrule(lr){2-2}
    \cmidrule(lr){3-3} 
    \cmidrule(lr){4-5}
    \cmidrule(lr){6-7}
    \cmidrule(lr){8-9}
    \cmidrule(lr){10-11}

    \multirow{2}{*}{supervised} & Combiner~\cite{Baldrati_2022_Combiner} & \textcolor{gray}{CVPR'22} & 36.36 & 58.00 & 31.63 & 56.67 & 38.19 & 62.42 & 35.39 & 59.03 \\

    \multirow{2}{*}{methods} & PL4CIR~\cite{zhao2022pl4cir}  & \textcolor{gray}{SIGIR'22} & 33.22 & 59.99 & 46.17 & 68.79 & 46.46 & 73.84 & 41.98 & 67.54 \\

    & Uncertainty retrieval~\cite{chen2024Uncertainty_retrieval}  & \textcolor{gray}{ICLR'24} & 32.61 & 61.34 & 33.23 & 62.55 & 41.40 & 72.51 & 35.75 & 65.47 \\

    \midrule 

    \multirow{9}{*}{ViT-L/14} & Image-only  & \textcolor{gray}{baseline} &  10.45 & 20.76 & \ \ 5.21 & 13.49 & \ \  8.01 & 18.05 & \ \  7.89 & 17.43 \\

    & Text-only  & \textcolor{gray}{baseline} & 20.26 & 34.10 & 15.12 & 33.71 & 21.98 & 39.98 & 19.12 & 35.93 \\

    & Image+Text  & \textcolor{gray}{baseline} & 19.14 & 32.63 & 14.38 & 31.09 & 20.50 & 36.26 & 18.01 & 33.33 \\

    & Pic2Word~\cite{Saito_2023_pic2word}  & \textcolor{gray}{CVPR'23} & 26.20 & 43.60 & 20.00 & 40.20 & 27.90 & 47.40 & 24.70 & 43.70 \\

    & SEARLE~\cite{Baldrati_2023_searle}  & \textcolor{gray}{ICCV'23} & 26.89 & 45.58 & 20.48 & 43.13 & 29.32 & 49.97 & 25.56 & 46.23 \\

    & LinCIR~\cite{Gu_2024_LinCIR}  & \textcolor{gray}{CVPR'24} & 29.10 & 46.81 & 20.92 & 42.44 & 28.81 & 50.18 & 26.28 & 46.49 \\

    & CIReVL~\cite{Shyamgopal_2024_CIReVL}  & \textcolor{gray}{ICLR'24} & \underline{29.49} & \underline{47.40} & \underline{24.79} & \underline{44.76} & \underline{31.36} & \bf 53.65 & \underline{28.55} & \underline{48.57} \\

    & CIG + LinCIR~\cite{Wang_2025_CIG}  & \textcolor{gray}{CVPR'25} & 28.90 & 47.25 & 21.12 & 43.88 & 29.78 & 50.54 & 26.60 & 47.22 \\
    
    & \bf Paracosm  & \textcolor{gray}{ours} & \bf 31.80 & \bf 49.51 & \bf 24.99 & \bf 47.45 & \bf 31.82 & \underline{52.83} & \bf 29.45 & \bf 49.93 \\

    \midrule 
    
    \multirow{7}{*}{ViT-G/14} & Pic2Word~\cite{Saito_2023_pic2word}  & \textcolor{gray}{CVPR'23} & 33.17 & 50.39 & 25.43 & 47.65 & 35.24 & 57.62 & 31.28 & 51.89 \\

    & SEARLE~\cite{Baldrati_2023_searle}  & \textcolor{gray}{ICCV'23} & 36.46 & 55.35 & 28.16 & 50.32 & 39.83 & 61.45 & 34.82 & 55.71 \\
    
    & CIReVL~\cite{Shyamgopal_2024_CIReVL}  & \textcolor{gray}{ICLR'24} & 33.71 & 51.42 & 27.07 & 49.53 & 35.80 & 56.14 & 32.19 & 52.36 \\
    & LDRE~\cite{yang_2024_ldre}  & \textcolor{gray}{SIGIR'24} & 35.94 & \bf{58.58} & 26.11 & 51.12 & 35.42 & 56.67 & 32.49 & 55.46 \\
    
    & AutoCIR~\cite{cheng2025autocir} & \textcolor{gray}{KDD'25} & 36.36 & 55.84 & 26.18 & 47.69 & 37.28 & 60.38 & 33.27 & 54.63\\

    & OSrCIR~\cite{Tang_2025_OSrCIR}  & \textcolor{gray}{CVPR'25} & \underline{38.65} & 54.71 & \underline{33.02} & \underline{54.78} & \underline{41.04} & \underline{61.83} & \underline{37.57} & \underline{57.11} \\
    
    & \bf Paracosm  & \textcolor{gray}{ours} & \bf 40.48	& \underline{57.80}	& \bf{33.17}	& \bf{55.18}	& \bf 42.58	& \bf{64.20}	& \bf 38.74	& \bf{59.06} \\

\bottomrule

\end{tabular}
}
\vspace{-5mm}
\label{tab:fashioniq_results} 
\end{table*}

{\bf Compared Methods.} 
We compare our Paracosm with representative and recent ZS-CIR methods, which can be categorized into training-based and training-free groups. We also compare baseline methods.
\begin{itemize}
\item {\em Baseline methods} include
``image-only'' which uses features of reference images for matching database images,
``text-only'' which uses modification text for cross-modality matching, and ``image+text'' which sums the image and text features for matching.

\item {\em Training-based methods} 
include {Pic2Word}~\cite{Saito_2023_pic2word},
{SEARLE}~\cite{Baldrati_2023_searle},
{LinCIR}~\cite{Gu_2024_LinCIR}, 
{CIG}~\cite{Wang_2025_CIG},
and {IP-CIR}~\cite{Li_2025_IPCIR}.
All these methods rely on textual features, e.g., training a textual inversion network to map reference images into pseudo-word tokens and merge them with textual tokens of modification texts.
Notably, 
IP-CIR~\cite{Li_2025_IPCIR} and CIG~\cite{Wang_2025_CIG} generate pseudo-target images for multimodal queries using a pretrained generative model.
Yet, they run along with other methods to ensure good CIR performance.

\item {\em Training-free methods.} 
{CIReVL}~\cite{Shyamgopal_2024_CIReVL} uses an LMM to generate a description of a given reference image, and then an LLM to modify this description based on the modification text. It uses the modified description for cross-modality retrieval in the image database.
{LDRE}~\cite{yang_2024_ldre} proposes to generate multiple diverse descriptions to improve CIR performance.
{AutoCIR}~\cite{cheng2025autocir} employs automatic multi-agent collaboration to iteratively refine retrieval queries through self-correction.
{CoTMR}~\cite{Sun_2025_cotmr} leverages an LMM to generate both global captions and fine-grained object-level descriptions combined with a multi-grained scoring that significantly improves the retrieval performance.
{OSrCIR}~\cite{Tang_2025_OSrCIR} carefully designs the prompt for an LMM with Reflective Chain-of-Thought to improve the output description quality for a multimodal query. 
\end{itemize}

\subsection{Benchmarking Results}
\label{ssec: benchmark_results}

{
\begin{table}[t]
\centering
\small
\caption{\small
\textbf{Efficiency comparisons.}
Computational costs are measured on the CIRCO benchmark.
We break down the costs
into (1) offline one-time database processing (w.r.t. wall-clock time, and
storage for synthetic images and features),
and (2) online per-query inference (w.r.t. GPU memory consumption and latency).
Notably, synthetic database images are generated only during pre-processing  
for feature extraction and do not need to be stored permanently; only 
compact features (0.38 GB) are retained. 
While Paracosm requires additional offline computation to mitigate synthetic-to-real domain gaps, it achieves state-of-the-art performance.
}
\vspace{-3mm}
\setlength{\tabcolsep}{0.82em}
\scalebox{0.85}
{
\begin{tabular}{l ccc r r ccccccc}
\toprule
Method &  \cellcolor{col11}{time} & \multicolumn{1}{c}{\cellcolor{col11}{\textcolor{lightgray}{img. stg.}}} & \multicolumn{1}{c}{\cellcolor{col11}{feat. stg.}}   &  \multicolumn{2}{c}{\cellcolor{col33}{inference}}    & CIRCO &  CIRR   &  FashionIQ   \\
 &  \cellcolor{col11}{hrs} &  \cellcolor{col11}{\textcolor{lightgray}{GB}}  &  \cellcolor{col11}{GB} & \cellcolor{col33}{GB} & \cellcolor{col33}{sec} &  mAP@5    &  R@1  &  R@10      \\
\midrule
AutoCIR~\cite{cheng2025autocir} & \cellcolor{col11}{0.1}      & \cellcolor{col11}{\textcolor{lightgray}{n/a}}    & \cellcolor{col11}{0.38}     & \cellcolor{col33}{2.3}    & \cellcolor{col33}{24}      & 24.05  & 31.81        & 30.68    \\
CoTMR~\cite{Sun_2025_cotmr}    & \cellcolor{col11}{0.1}      & \cellcolor{col11}{\textcolor{lightgray}{n/a}}    & \cellcolor{col11}{0.38}     & \cellcolor{col33}{70.2}     & \cellcolor{col33}{1}   & 27.61 & 35.02        & 35.05   \\
OSrCIR~\cite{Tang_2025_OSrCIR}    & \cellcolor{col11}{0.1}      & \cellcolor{col11}{\textcolor{lightgray}{n/a}}    & \cellcolor{col11}{0.38}     & \cellcolor{col33}{2.3}    & \cellcolor{col33}{3}  & 23.87 & 29.45          & 33.26     \\
\textbf{Paracosm}  &  \cellcolor{col11}{12.9}  &  \cellcolor{col11}{\textcolor{lightgray}{41.4}}  &  \cellcolor{col11}{0.38}  &   \cellcolor{col33}{2.7}   &  \cellcolor{col33}{14}  &  37.40 &  38.24   &  36.45 \\
\bottomrule
\end{tabular}
}
\vspace{-5mm}
\label{tab:efficiency_comparisons_tab}
\end{table}
}

{\bf Quantitative Results.} 
Table~\ref{tab:cirr_circo_results} and \ref{tab:fashioniq_results}
display benchmarking results of all the compared methods across different VLM backbones, with respect to different metrics on all the test sets.
Convincingly, Paracosm significantly outperforms all compared methods.
Generally, with more powerful VLM models, e.g., ViT-G/14 $>$ ViT-L/14, methods yield better CIR performance.
Paracosm even rivals recent supervised learning CIR methods~\cite{zhao2022pl4cir, chen2024Uncertainty_retrieval}.

{\bf Qualitative Results.} 
\cref{fig:fig_generated_img_has_more_visual_informantion} visually compares the retrieved images by Paracosm and the recent method OSrCIR~\cite{Tang_2025_OSrCIR}, along with their generated mental images and descriptions, respectively.
Clearly, generated descriptions by OSrCIR insufficiently represent the given multimodal queries, whereas mental images by Paracosm preserve and capture rich visual details pertaining to the multimodal queries.
This helps explain why Paracosm outperforms OSrCIR.

\begin{table*}[t]
\centering
\small
\caption{\small
\textbf{Rigorous comparisons between using OpenCLIP vs. CLIP backbones.} 
We copy the results of OSrCIR~\cite{Tang_2025_OSrCIR} from its original paper (marked in {\setlength{\fboxsep}{1pt}\colorbox{lightlightgrey}{grey shade}}), which reports that it uses the CLIP ViT-B/32 backbone and GPT-4o in experiments.
We reproduce it in the same setting but achieve notably worse performance.
We also adapt OSrCIR by using the OpenCLIP backbone with either GPT-4o or Qwen2.5-VL.
Notably, the results achieved with OpenCLIP (in {\setlength{\fboxsep}{1pt}\colorbox{col33}{blue shade}}) match what are reported in OSrCIR~\cite{Tang_2025_OSrCIR}.
This suggests that the backbone used by OSrCIR is likely to be OpenCLIP, instead of CLIP.
Our study helps answer questions regarding the performance gaps in the community,
as seen in the Issues of the OSrCIR's official GitHub repository \cite{OSrCIR_issue}.
Refer to the main text for more analyses.
}
\vspace{-3mm}
\setlength{\tabcolsep}{0.75em}
\scalebox{0.75}
{
    \begin{tabular}{lllccccccccccccc}
    \toprule
    \multicolumn{4}{c}{} 
    & \multicolumn{4}{c}{CIRCO} & \multicolumn{2}{c}{Fashion IQ}\\
    \cmidrule(lr){5-8}
    \cmidrule(lr){9-10}
    & & &
    & \multicolumn{4}{c}{mAP@k} 
    & \multicolumn{2}{c}{Average} 

    \\
   
     \textbf{Backbone} & \textbf{Method} & \textbf{LMMs}  &  \textcolor{gray}{venue\&year}  &  k=5 & k=10 &  k=25 & k=50 & R@10 &  R@50\\
    \cmidrule(lr){1-1}
    \cmidrule(lr){2-2}
    \cmidrule(lr){3-3} 
    \cmidrule(lr){4-4}
    \cmidrule(lr){5-8}
    \cmidrule(lr){9-10}

    \multirow{3}{*}{CLIP} & \cellcolor{lightlightgrey}OSrCIR  & \cellcolor{lightlightgrey}GPT-4o  & \cellcolor{lightlightgrey}\textcolor{gray}{CVPR'25} & \cellcolor{lightlightgrey}{\textcolor{darkgray}{18.04}} & \cellcolor{lightlightgrey}{\textcolor{darkgray}{19.17}} & \cellcolor{lightlightgrey}{\textcolor{darkgray}{20.94}} & \cellcolor{lightlightgrey}{\textcolor{darkgray}{21.85}}  & \cellcolor{lightlightgrey}{\textcolor{darkgray}{32.34}} & \cellcolor{lightlightgrey}{\textcolor{darkgray}{53.40}} \\
    
    \multirow{3}{*}{ViT-B/32} & OSrCIR & GPT-4o  & \textcolor{gray}{reproduced} & 14.91 & 15.34 & 16.79 & 17.60 & 19.15 & 36.88 \\
									
    & OSrCIR  & Qwen2.5-VL  & \textcolor{gray}{reproduced} & 11.84 & 12.39 & 13.53 & 14.27 & 18.75 & 36.05 \\

    & \bf Paracosm & Qwen2.5-VL & \textcolor{gray}{ours} & 26.10 & 27.02 & 29.29 & 30.45 & 26.14 & 46.39 \\
    
    \midrule 

    \multirow{3}{*}{CLIP} & \cellcolor{lightlightgrey}OSrCIR  & \cellcolor{lightlightgrey}GPT-4o  & \cellcolor{lightlightgrey}\textcolor{gray}{CVPR'25} & \cellcolor{lightlightgrey}{\textcolor{darkgray}{23.87}} & \cellcolor{lightlightgrey}{\textcolor{darkgray}{25.33}} & \cellcolor{lightlightgrey}{\textcolor{darkgray}{27.84}} & \cellcolor{lightlightgrey}{\textcolor{darkgray}{28.97}} & \cellcolor{lightlightgrey}{\textcolor{darkgray}{33.26}} & \cellcolor{lightlightgrey}{\textcolor{darkgray}{54.37}}\\

    \multirow{3}{*}{ViT-L/14} & OSrCIR  & GPT-4o  & \textcolor{gray}{reproduced} & 17.70 & 18.56 & 20.52 & 21.47 & 23.01 & 41.39 \\
									
    & OSrCIR  & Qwen2.5-VL  & \textcolor{gray}{reproduced} & 14.06 & 14.93 & 16.65 & 17.50 & 20.95 & 39.49\\
    
    & \bf Paracosm & Qwen2.5-VL & \textcolor{gray}{ours} & 30.24 & 31.51 & 34.29 & 35.42 & 28.76 & 49.32 \\

    \midrule
    
    \multirow{2}{*}{OpenCLIP} & \cellcolor{col33}OSrCIR  & \cellcolor{col33}GPT-4o  & \cellcolor{col33}\textcolor{gray}{reproduced} & \cellcolor{col33}{18.77} & \cellcolor{col33}{19.33} & \cellcolor{col33}{21.04} & \cellcolor{col33}{22.02} & \cellcolor{col33}{31.41} & \cellcolor{col33}{52.26}\\
									
    \multirow{2}{*}{ViT-B/32} & OSrCIR  & Qwen2.5-VL  & \textcolor{gray}{reproduced} & 14.47 & 15.23 & 16.70 & 17.55 & 26.54 & 47.35\\

    & \bf Paracosm & Qwen2.5-VL & \textcolor{gray}{ours}    & 31.31 & 32.48 & 35.19 & 36.33 & 35.14 & 56.04\\

    \midrule 
    
    \multirow{2}{*}{OpenCLIP} & \cellcolor{col33}OSrCIR  & \cellcolor{col33}GPT-4o  & \cellcolor{col33}\textcolor{gray}{reproduced}  & \cellcolor{col33}{22.75} & \cellcolor{col33}{23.51} & \cellcolor{col33}{25.73} & \cellcolor{col33}{26.78} & \cellcolor{col33}{31.55} & \cellcolor{col33}{51.90} \\
									
    \multirow{2}{*}{ViT-L/14} & OSrCIR  & Qwen2.5-VL  & \textcolor{gray}{reproduced} & 17.88 & 18.72 & 20.51 & 21.39 & 27.84 & 47.16 \\

    & \bf Paracosm & Qwen2.5-VL & \textcolor{gray}{ours} & 37.40 & 38.64 & 41.57 & 42.73 & 36.45 & 57.60 \\
    
\bottomrule

\end{tabular}
}
\vspace{-0mm}
\label{tab:osrcir_suppl} 
\end{table*}

{\bf Efficiency Comparisons.}
For a comprehensive analysis,
we evaluate the computational costs of Paracosm and some recent zero-shot methods under the same experimental conditions in~\Cref{tab:efficiency_comparisons_tab}.
To ensure a fair comparison, 
all methods are evaluated using the same VLM backbone (OpenCLIP ViT-L/14~\cite{ilharco_gabriel_2021_openclip}) 
for feature extraction and matching and are run on NVIDIA A100 GPUs.
Regarding LMM backbones, AutoCIR~\cite{cheng2025autocir} and OSrCIR~\cite{Tang_2025_OSrCIR} use GPT~\cite{openai2024gpt4technicalreport}, whereas CoTMR~\cite{Sun_2025_cotmr} and our Paracosm use Qwen~\cite{bai2025qwen2}.
Computational costs are measured on the CIRCO benchmark \cite{Baldrati_2023_CIRCO},
which comprises over 123K database images.
The results show that 
(1) Paracosm incurs substantial wall time hours to process database images,
(2) but its inference efficiency is comparable to AutoCIR,
and (3) importantly, Paracosm significantly outperforms these methods!

{\bf A Warning.}
It is worth noting that some results of recent studies are not reproducible probably because of misreported backbones in the papers.
We find that using OpenCLIP \cite{ilharco_gabriel_2021_openclip} achieves better ZS-CIR performance than  CLIP \cite{radford2021clip}, which is reported to be used for experiments in some papers such as OSrCIR~\cite{Tang_2025_OSrCIR}.
To rigorously analyze this issue, 
in  \Cref{tab:osrcir_suppl}, we copy the results reported by OSrCIR and reproduce it in the same setting and under different settings.
Clearly, the results achieved by using OpenCLIP in our implementation match what are reported in \cite{Tang_2025_OSrCIR}, which, however, reports using CLIP.
Our analysis supports the doubts in the community,
as seen in the Issues of their official GitHub repositories \cite{OSrCIR_issue}. 
Moreover, 
results in \Cref{tab:osrcir_suppl} show that using GPT-4o achieves better performance than Qwen2.5-VL for OSrCIR,
while our Paracosm, which uses Qwen2.5-VL, outperforms OSrCIR regardless of using Qwen2.5-VL or GPT-4o.

\subsection{Analyses and Ablation Study}
\label{ssec:ablation}


{\bf Reference image editing vs. Text-to-Image generation.}
When generating mental images for multimodal queries, our Paracosm leverages the LMM 
to directly edit the reference images based on modification texts. 
In comparison, another workaround is T2I generation,
as done in the recent work \cite{Li_2025_IPCIR}:
generating descriptions for reference images using an LMM, then modifying the descriptions based on modification texts using an LLM, and lastly generating mental images based on the modified descriptions.
\Cref{tab:further_analysis} compares them, clearly showing that Paracosm's design choice leads to better performance.

{\bf Different image generator.}
We compare the results of Paracosm using Qwen-Image-Edit~\cite{wu2025qwenimagetechnicalreport} and another open-source image generator, LongCat-Image-Edit~\cite{team2025longcat}, on the CIRR test set without tuning hyperparameters or prompts.
The results in \Cref{tab:further_analysis} show stable performance of Paracosm with different generators.

\begin{figure}[t]
\centering
\begin{minipage}[c]{0.48\textwidth}
\includegraphics[width=\textwidth]{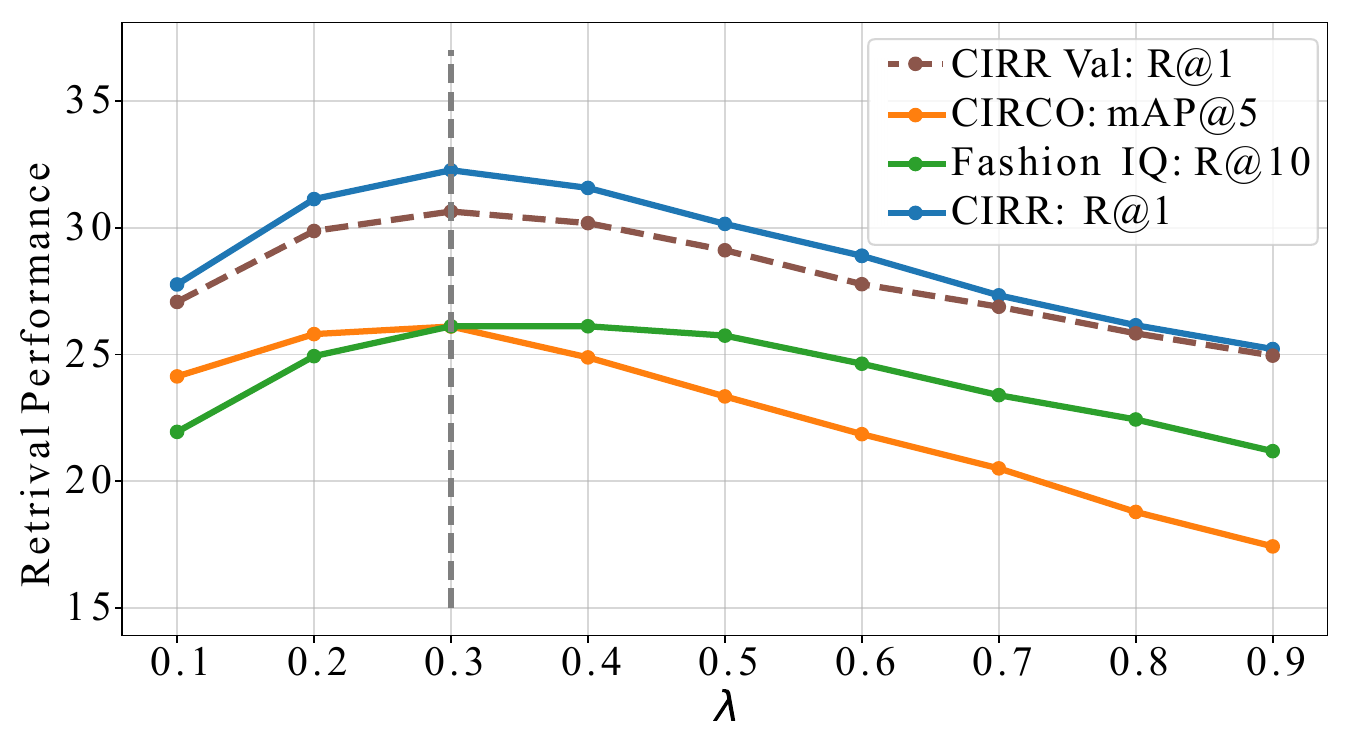}
\end{minipage}
\hfill
\begin{minipage}[c]{0.48\textwidth}
\caption{\small
  {\bf Analysis of $\lambda$} which controls the importance of incorporating modification text in Eq.~\ref{eq:features}.
  We set $\lambda=0.3$ based on the results on the CIRR validation set.
  Interestingly, on all datasets, setting $\lambda=0.3$ consistently yields the highest numeric metrics reported for all the datasets.
  }
\label{fig:lambda}
\end{minipage}
\vspace{-6mm}
\end{figure}

{\bf Ablation Study.}
We ablate core components in Paracosm with the CLIP ViT-B/32 VLM on CIRR and CIRCO. 
\Cref{tab:cirr_circo_results_ablation} summarizes key results; supplementary \Cref{sec: Detail_results_suppl} provides comprehensive details.
\begin{itemize}
    \item {\em Incorporating mental images boosts performance.}
        While existing training-free methods generate descriptions ${\mathbf t}_{query}$, aka pseudo-target descriptions \cite{Tang_2025_OSrCIR}, Paracosm further incorporates its generated mental images $\I_{mental}$.
        This results in clear performance gains.
    
    \item {\em Incorporating synthetic counterparts of database images boosts performance.}        
        As $\I_{mental}$ has synthetic-to-real domain gaps with database images, Paracosm further generates synthetic counterparts $\I_{syn}^i$ for them and uses such for matching (Eq.~\ref{eq:features}). 
        Clearly, doing this significantly boosts the final performance.
        
    \item {\em Incorporating modification texts boosts performance.}
        Intuitively, modification texts ${\mathbf t}_{mod}$ contain crucial keywords that can be leveraged for CIR.
        Indeed, our ablation results demonstrate that incorporating modification texts greatly boosts CIR performance. This observation aligns with what is reported in prior works~\cite{Baldrati_2023_searle, Gu_2024_LinCIR, Wang_2025_CIG}.

\end{itemize}

\begin{table*}[t]
\centering
\small
\caption{\small
\textbf{Ablation study.} 
We ablate key components of Paracosm with the CLIP ViT-B/32 VLM on the CIRCO and CIRR test sets.
Per Eq.~\ref{eq:features}, 
we focus on constructing features for {\setlength{\fboxsep}{1pt}\colorbox{col11}{multimodal queries}} and {\setlength{\fboxsep}{1pt}\colorbox{col33}{database images}}.
$\I_{mental}$, ${\mathbf t}_{query}$, and ${\mathbf t}_{mod}$ indicate the generated mental image, description of the target image based on the query, and modification text, respectively.
$\I^i$ and $\I_{syn}^i$ represent database images and their synthetic counterparts.
Best results are highlighted in \textbf{bold}. 
Clearly, incorporating $\I_{mental}$, $\I_{syn}$, and ${\mathbf t}_{mod}$ significantly boosts CIR performance.
} 
\vspace{-3mm}
\setlength{\tabcolsep}{0.6em}
\scalebox{0.6}
{
    \begin{tabular}{lllcccccccccccccc}
    \toprule
    \multicolumn{3}{c}{\cellcolor{col11}{multimodal query}} & \multicolumn{2}{c}{\cellcolor{col33}database images}
    & \multicolumn{7}{c}{CIRR} & \multicolumn{4}{c}{CIRCO}\\
    \cmidrule(lr){1-3}
    \cmidrule(lr){4-5}
    \cmidrule(lr){6-12}
    \cmidrule(lr){13-16}
    
    \cellcolor{col11}  & \cellcolor{col11}   & \cellcolor{col11}  & \cellcolor{col33} & \cellcolor{col33} 
    & \multicolumn{4}{c}{Recall@k} 
    & \multicolumn{3}{c}{Recall$_{\text{Subset}}$@k} 
    & \multicolumn{4}{c}{mAP@k}  

    \\
   
     \cellcolor{col11}\textbf{${\mathbf t}_{query}$} & \cellcolor{col11}\textbf{$\I_{mental}$} & \cellcolor{col11}\textbf{${\mathbf t}_{mod}$} & \cellcolor{col33}\textbf{$\I^i$}  & \cellcolor{col33}\textbf{$\I_{syn}$} & k=1 &  k=5  & k=10 &  k=50  & k=1 &  k=2  & k=3 &  k=5 & k=10 &  k=25 & k=50 \\
    \cmidrule(lr){1-3}
    \cmidrule(lr){4-5}
    \cmidrule(lr){6-9}
    \cmidrule(lr){10-12}
    \cmidrule(lr){13-16}

     \cellcolor{col11}\checkmark &  \cellcolor{col11}  &  \cellcolor{col11} & \cellcolor{col33}\checkmark & \cellcolor{col33} &  17.21 & 43.49 & 55.90 & 81.16 & 52.00 & 72.31 & 84.92 & 14.91 & 15.34 & 16.79 & 17.60 \\

     \cellcolor{col11}\checkmark &  \cellcolor{col11}\checkmark & \cellcolor{col11} & \cellcolor{col33}\checkmark & \cellcolor{col33} &  18.80 & 44.96 & 58.84 & 82.65 & 50.39 & 72.80 & 83.93 & 13.71 & 13.89 & 15.19 & 15.92 \\

     \cellcolor{col11}\checkmark  &  \cellcolor{col11}\checkmark &  \cellcolor{col11}\checkmark & \cellcolor{col33}\checkmark & \cellcolor{col33} &  27.93 & 57.11 & 70.29 & 90.31 & 61.88 & 80.70 & 90.70 & 18.29 & 18.69 & 20.45 & 21.27 \\

    \midrule

    \cellcolor{col11}\checkmark  &  \cellcolor{col11} &  \cellcolor{col11} & \cellcolor{col33} & \cellcolor{col33}\checkmark  & 17.21 & 43.93 & 56.29 & 80.10 & 50.94 & 71.81 & 84.53 & 13.58 & 13.92 & 15.44 & 16.30 \\

     \cellcolor{col11}\checkmark &  \cellcolor{col11}\checkmark & \cellcolor{col11} & \cellcolor{col33} & \cellcolor{col33}\checkmark  & 19.95 & 46.29 & 59.98 & 82.34 & 51.42 & 72.17 & 84.80 & 14.07 & 14.43 & 15.76 & 16.42 \\

     \cellcolor{col11}\checkmark  &  \cellcolor{col11}\checkmark &  \cellcolor{col11}\checkmark & \cellcolor{col33} & \cellcolor{col33}\checkmark  & 25.95 & 54.24 & 67.16 & 88.84 & 60.80 & 79.71 & 89.74 & 17.43 & 17.98 & 19.62 & 20.46 \\

     \midrule

     \cellcolor{col11}\checkmark & \cellcolor{col11}  &   \cellcolor{col11} & \cellcolor{col33}\checkmark & \cellcolor{col33}\checkmark & 22.46 & 50.99 & 63.59 & 84.58 & 53.88 & 74.80 & 86.27 & 16.72 & 17.41 & 19.19 & 20.11 \\

     \cellcolor{col11}\checkmark &  \cellcolor{col11}\checkmark & \cellcolor{col11} & \cellcolor{col33}\checkmark & \cellcolor{col33}\checkmark & 24.60 & 52.68 & 65.86 & 86.53 & 54.84 & 74.92 & 86.82 & 16.57 & 17.10 & 18.59 & 19.30 \\

     \cellcolor{col11}\checkmark  &  \cellcolor{col11}\checkmark &  \cellcolor{col11}\checkmark & \cellcolor{col33}\checkmark & \cellcolor{col33}\checkmark & \textbf{32.27} & \textbf{62.60} & \textbf{75.16} & \textbf{92.60} & \textbf{65.16} & \textbf{83.25} & \textbf{92.34} & \textbf{26.10} & \textbf{27.02} & \textbf{29.29} & \textbf{30.45} \\

\bottomrule

\end{tabular}
}
\vspace{-5mm}
\label{tab:cirr_circo_results_ablation} 
\end{table*}

\section{Impacts and Limitations}
\label{sec:discussions}

{\bf Societal Impacts.}
CIR is an important component of search services in many real-world applications, such as e-commerce and the fashion industry.
The proposed Paracosm, by extensively leveraging LMMs, greatly advances the CIR area, as demonstrated by its state-of-the-art performance on standard benchmark datasets.
It is worth noting that LMMs' large-scale pretraining datasets might contain poisonous, offensive, or biased data. This could cause the LMMs to generate inaccurate, biased, or even harmful text descriptions and synthetic images.
Moreover, 
Paracosm does not have an alerting mechanism for inappropriate multimodal queries, which may contain malicious modification texts or intend to retrieve inappropriate target images.
If forced to retrieve target images based on such queries,
the intermediate mental images and retrieval outputs may lead to negative societal impacts.

\begin{figure}[t]
\centering
\includegraphics[width=\textwidth]{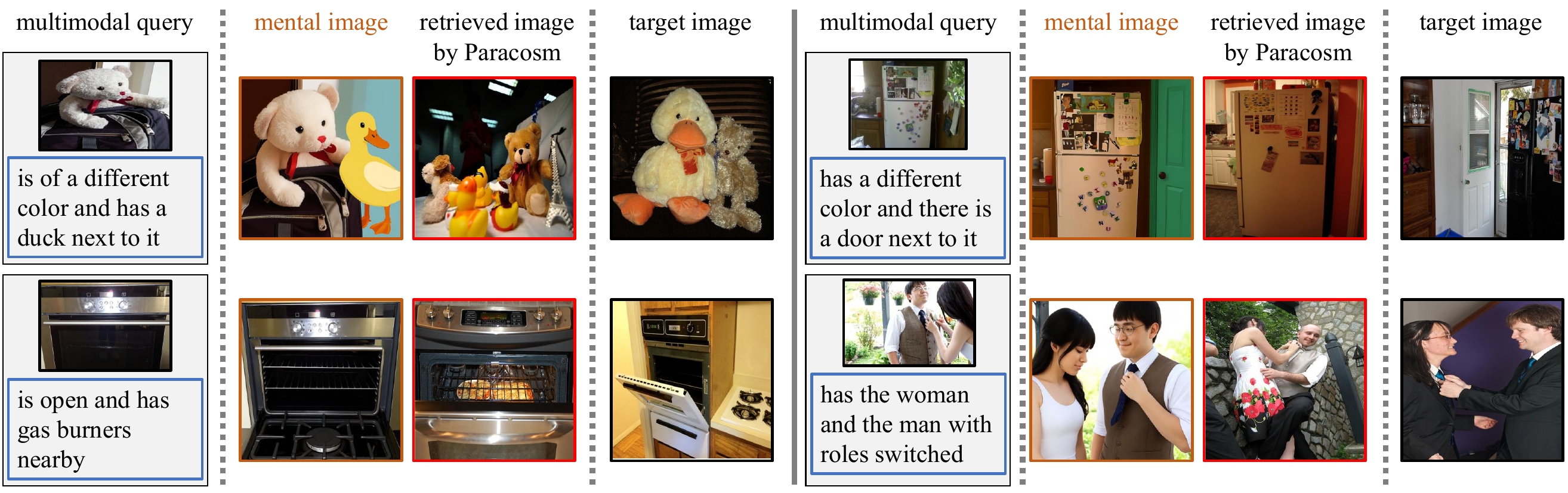}
\vspace{-6.5mm}
\caption{\small
  {\bf Failure cases} on the CIRCO dataset.
  Paracosm can fail due to limitations of generative models that can generate implausible and counterfactual mental images. 
  Four examples, respectively, demonstrate different failures of Paracosm. 
  (1) It incorrectly generates a cartoon-style duck, making it fail to return the correct database image, which captures a plush toy duck.
  (2) It generates a counterfactual oven that has gas burners on its door, making it incorrectly retrieve an image that captures a burning oven.
  (3) It fails to edit the door color of the specified refrigerator and hence fails to return the correct target image. 
  (4) It fails to comprehend the multimodal query, resulting in an incorrect mental image and hence failing to return the target image.
  }
\label{fig:generated_img_not_good}
\vspace{-3mm}
\end{figure}

{\bf Limitations and Future Work.}
Paracosm relies on the quality of LMM-generated mental images for queries and synthetic counterparts of database images, as well as text descriptions for them.
This means that the performance of Paracosm is inherently contingent upon the generative 
capabilities of the underlying LMMs. 
These generated images might look visually plausible at first sight but often lack factual fidelity and fine-grained details.
For instance, in the first row of~\Cref{fig:generated_img_not_good}, the mental image contains a cartoon-style duck, which does not correspond to any actual items in the real world and causes retrieval to fail.
Future work can focus on improving methods for image generation or editing and developing methods to handle erroneously generated visuals.
Moreover,
to achieve state-of-the-art performance, Paracosm uses slightly different prompt templates in the benchmarking datasets.
This is primarily due to variations in the format and specificity of modification texts across datasets, as shown in \Cref{fig:prompt}.
For example, a modification text can contain a relative caption and a
shared concept(ref. CIRCO).
To overcome this, future work could develop adaptive prompt generation methods that leverage LMMs to automatically construct optimal prompts for diverse types of multimodal 
queries, reducing the need for manual tuning and enhancing robustness across 
varying dataset distributions.

\section{Conclusions}
We present Paracosm, a novel and rather simple training-free zero-shot CIR method.
Different from most existing ZS-CIR methods, which rely on generated text descriptions of multimodal queries,
Paracosm leverages LMMs to generate mental images for multimodal queries, offering rich visual information that is beneficial to CIR.
Moreover, as the synthetic mental images have synthetic-to-real domain gaps with real database images,
it further generates synthetic counterparts of database images.
By incorporating the generated visuals as well as the generated descriptions, it achieves state-of-the-art performance on popular benchmarking datasets.
Paracosm significantly outperforms existing ZS-CIR methods and even rivals supervised learning approaches.

\section*{Acknowledgments}
This work was supported by the Science and Technology Development Fund of Macau (0058/2025/RIA2, 0067/2024/ITP2), the University of Macau (SRG2023-00044-FST), the Institute of Collaborative Innovation, and the CK Foundation. Authors thank Hanxin Wang, Di Wu, and Nan Deng for helpful discussions.

\bibliographystyle{splncs04}
\bibliography{main}
\clearpage
\setcounter{page}{1}

{
\begin{center}
\Large
\textbf{Generating a Paracosm for Training-Free Zero-Shot Composed Image Retrieval}\\(Supplementary Material)
\end{center}
}

\renewcommand{\thesection}{\Alph{section}}
\renewcommand{\theHsection}{\Alph{section}}
\setcounter{section}{0} 



This supplementary material provides detailed results and comprehensive analyses to complement the main paper. Here is the outline of the document: 

\begin{itemize}
\item {\bf Section \ref{sec:datasets_suppl}} provides additional details on benchmark datasets. 


\item {\bf Section \ref{sec:img_suppl}} presents qualitative analysis of mental images for queries and synthetic counterparts for database images.

\item {\bf Section \ref{sec:further_analysis_suppl}} presents additional analyses and ablation studies.

\item {\bf Section \ref{sec:open_source_code_suppl}} introduces the demo code included in this supplementary material.

\item {\bf Section \ref{sec: Detail_results_suppl}} provides detailed quantitative results for each benchmark dataset.

\end{itemize}

\section{Additional Details of Benchmark Datasets}
\label{sec:datasets_suppl}

In this section, we provide additional details on the benchmark datasets used in our experiments, namely CIRR~\cite{Liu_2021_CIRR}, CIRCO~\cite{Baldrati_2023_CIRCO}, and Fashion IQ~\cite{Wu_2021_FashionIQ}.
We summarize these datasets used in \Cref{tab:dataset_supply}.
\begin{itemize}

    \item The Compose Image Retrieval on Real-life images (CIRR) dataset constructs negative samples by mining visually similar images from NLVR2~\cite{suhr2019corpus}.
    To mitigate the false negative issue, CIRR organizes visually similar images into subsets. Since each subset contains images that are visually close to the target, evaluating retrieval within this constrained subset places greater demands on the model's discriminative capability. 

    \item Composed Image Retrieval on Common Objects in context (CIRCO) is based on open-domain real-world images~\cite{lin2014microsoft}. It is the first dataset for CIR with multiple ground-truth targets per query and fine-grained annotations. 
    \item Fashion Interactive Queries (Fashion IQ) contains fashion images across three categories (Shirt, Dress, Toptee) from Amazon.com. Following prior work \cite{Saito_2023_pic2word, Baldrati_2023_searle, cheng2025autocir, yang_2024_ldre, Li_2025_IPCIR}, we evaluate on the validation set, as the test set is not publicly released.
\end{itemize}

\begin{figure}[t]
\centering
\includegraphics[width=1\linewidth, clip=true,trim = 0mm 0mm 0mm 0mm]{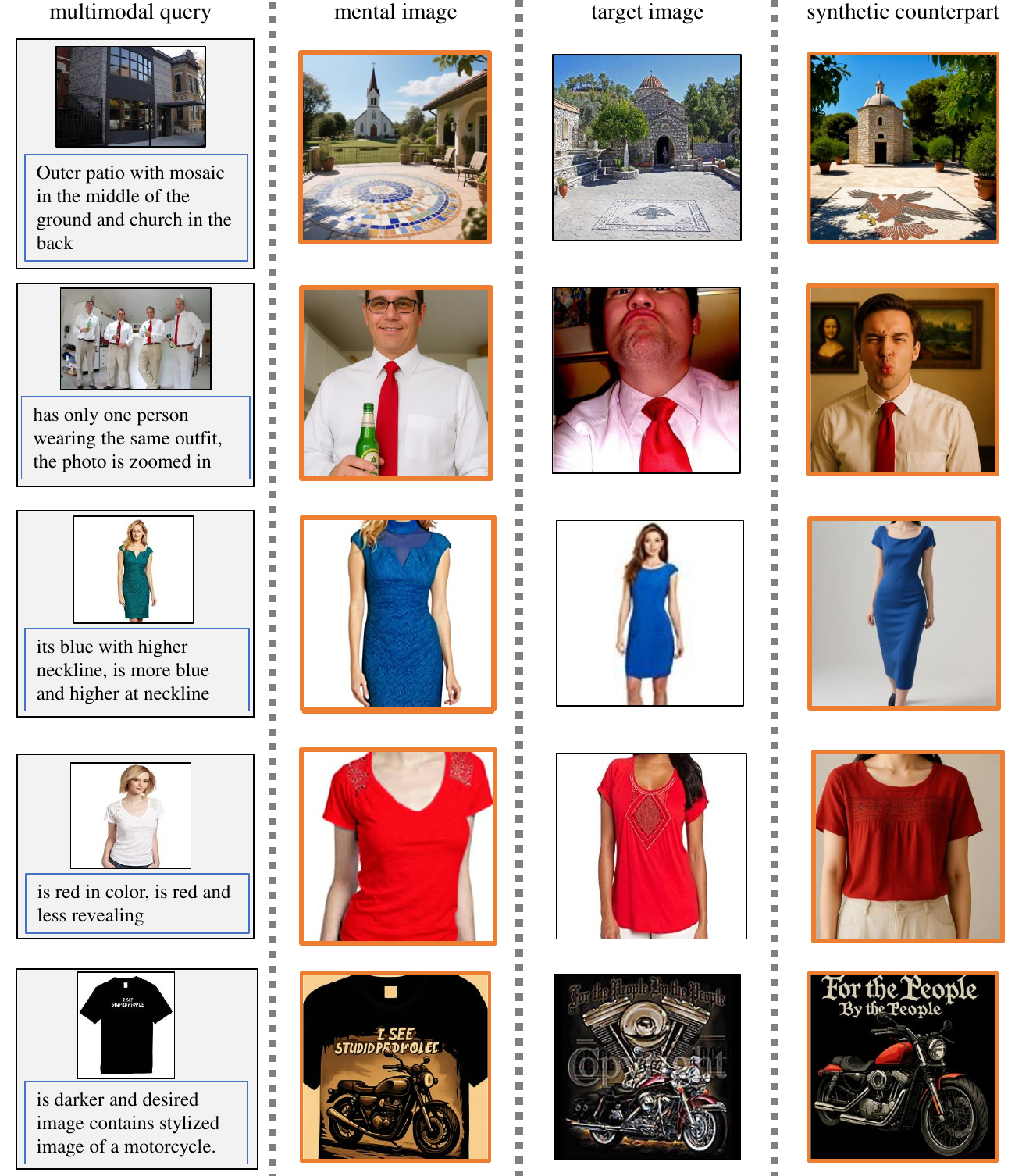}
\vspace{-6mm}
\caption{\small
  \textbf{Additional qualitative examples of Paracosm} from CIRR \cite{Liu_2021_CIRR}, CIRCO \cite{Baldrati_2023_CIRCO}, and Fashion IQ \cite{Wu_2021_FashionIQ}. The first column presents the reference image paired with its modification text. The second column displays the mental images generated for the multimodal queries. The third column shows the ground-truth target images. The fourth column contains synthetic counterparts generated from the detailed textual descriptions of database images. 
  }
\label{fig:fig_generated_img_four_datasets_suppl}
\vspace{-7mm}
\end{figure}

\section{Qualitative Analysis of Mental and Synthetic Counterparts}
\label{sec:img_suppl}
To gain deeper insight into the behavior of our Paracosm, we visualize the generated mental images for multimodal queries and synthetic counterparts for database images across datasets.
As shown in \Cref{fig:fig_generated_img_four_datasets_suppl}, we leverage Qwen-Image-Edit~\cite{wu2025qwenimagetechnicalreport} to generate mental images from reference images, conditioned on the provided modification texts. 
Since the generated mental images are synthetic, they exhibit synthetic-to-real domain gaps relative to real database images.
To address this, we generate a synthetic counterpart for each database image based on its detailed textual description, which is generated by an LMM.
These synthetic counterparts preserve fine-grained visual attributes of the original images while aligning both queries and database images into a unified synthetic space,
thereby mitigating the domain gap and enabling more effective retrieval.

\begin{table}[t]
\centering
\begin{minipage}[c]{0.48\textwidth}
\setlength{\tabcolsep}{0.2em}
\scalebox{0.88}
{
    \begin{tabular}{llrrcccccc}
    \toprule
    \multicolumn{1}{l}{Dataset} &  \multicolumn{1}{l}{subset} 
    & \multicolumn{1}{c}{\# query} & \multicolumn{1}{c}{\# database}  \\
    \cmidrule(lr){1-4}

    CIRR~\cite{Liu_2021_CIRR} & -- & 4,148 & 2,315 \\
    \midrule
    CIRCO~\cite{Baldrati_2023_CIRCO}  & -- & 800 & 123,403 \\
    \midrule
    \multirow{3}{*}{Fashion IQ~\cite{Wu_2021_FashionIQ}}   & Shirt & 2,038 & 6,346 \\

      & Dress & 2,017 & 3,817 \\

      & Toptee & 1,961 & 5,373 \\



    
\bottomrule

\end{tabular}
}
\end{minipage}
\hfill
\begin{minipage}[c]{0.48\textwidth}
\caption{\small
\textbf{Statistics of datasets for ZS-CIR.} 
We list query counts in the test/validation sets and the numbers of database images for each benchmark. 
Since Paracosm is a training-free ZS-CIR method, we only utilize the officially released evaluation splits: 
test sets for CIRR and CIRCO, and the validation set for Fashion IQ. 
}
\vspace{-5mm}
\label{tab:dataset_supply} 
\end{minipage}
\end{table}

\begin{figure*}[t]
  \centering
  \includegraphics[width=1\linewidth, page=1, clip=true,trim = 0mm 0mm 0mm 0mm]{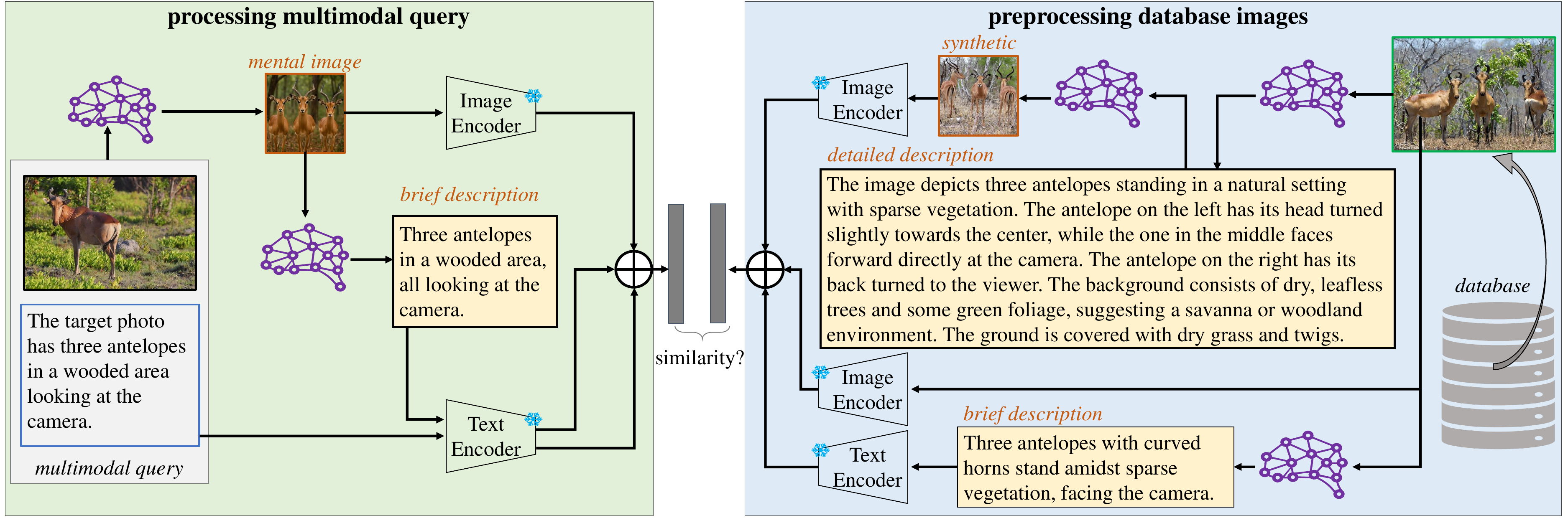}
  \hfill
  \vspace{-5mm}
  \caption{\small
  {\bf Extended flowchart of Paracosm incorporating brief database image descriptions (w/ $\mathbf{t}^i_{brief}$).} This variant incorporates brief textual descriptions of database images into the retrieval process. This version supports the study in \Cref{tab:further_analysis_supply}, which evaluates the impact of brief database image descriptions on performance.
} 
\label{fig:framework_suppl}
\vspace{-0mm}
\end{figure*}

\section{Further Analyses and Ablation Study}
\label{sec:further_analysis_suppl}
We further analyze the impact of different LMMs for generating textual descriptions of mental images. Specifically, we replace Qwen2.5-VL-7B-Instruct~\cite{qwen2.5-VL} used in Paracosm with GPT-4o, the LMM employed by OSrCIR~\cite{Tang_2025_OSrCIR}.
As shown in \Cref{tab:further_analysis_supply}, switching to GPT-4o yields only marginal improvements in retrieval performance.
The empirical results in \Cref{tab:robussness_analysis_supply} confirm the exceptional robustness of Paracosm.
Notably, the best performance across all datasets is achieved when $\lambda$ = 0.3 (\cref{fig:lambda}), indicating that retrieval performance is relatively insensitive to the choice of LMM for mental image description generation. 
This indicates that the core strength of Paracosm lies not in the scale of LMMs but in the unified synthetic space (``paracosm'') constructed by our method.

Additionally, we evaluate the effect of incorporating brief textual descriptions ($\mathbf{t}^i_{brief}$) for database images into Paracosm, as illustrated in \cref{fig:framework_suppl}.
However, as shown in \Cref{tab:further_analysis_supply}, this strategy yields performance gains only on the CIRR and CIRCO datasets, with negligible improvements on Fashion IQ. 
Consequently, we opt to exclude $\mathbf{t}^i_{brief}$ from our final pipeline to maintain 
simplicity and efficiency. 
Beyond the ablation study in \Cref{tab:cirr_circo_results_ablation}, we further analyze an additional variant in \Cref{tab:cirr_circo_results_ablation_suppl} that incorporates detailed textual descriptions of database images into the matching process. However, this variant yields degraded performance compared to our final design. 
We attribute this decline to the granularity mismatch between the detailed descriptions of database images (which capture comprehensive visual attributes) and the brief descriptions of mental images (which focus on core semantic content).

\begin{figure}[t]
\centering
\begin{minipage}[c]{0.48\textwidth}
\includegraphics[width=\textwidth]{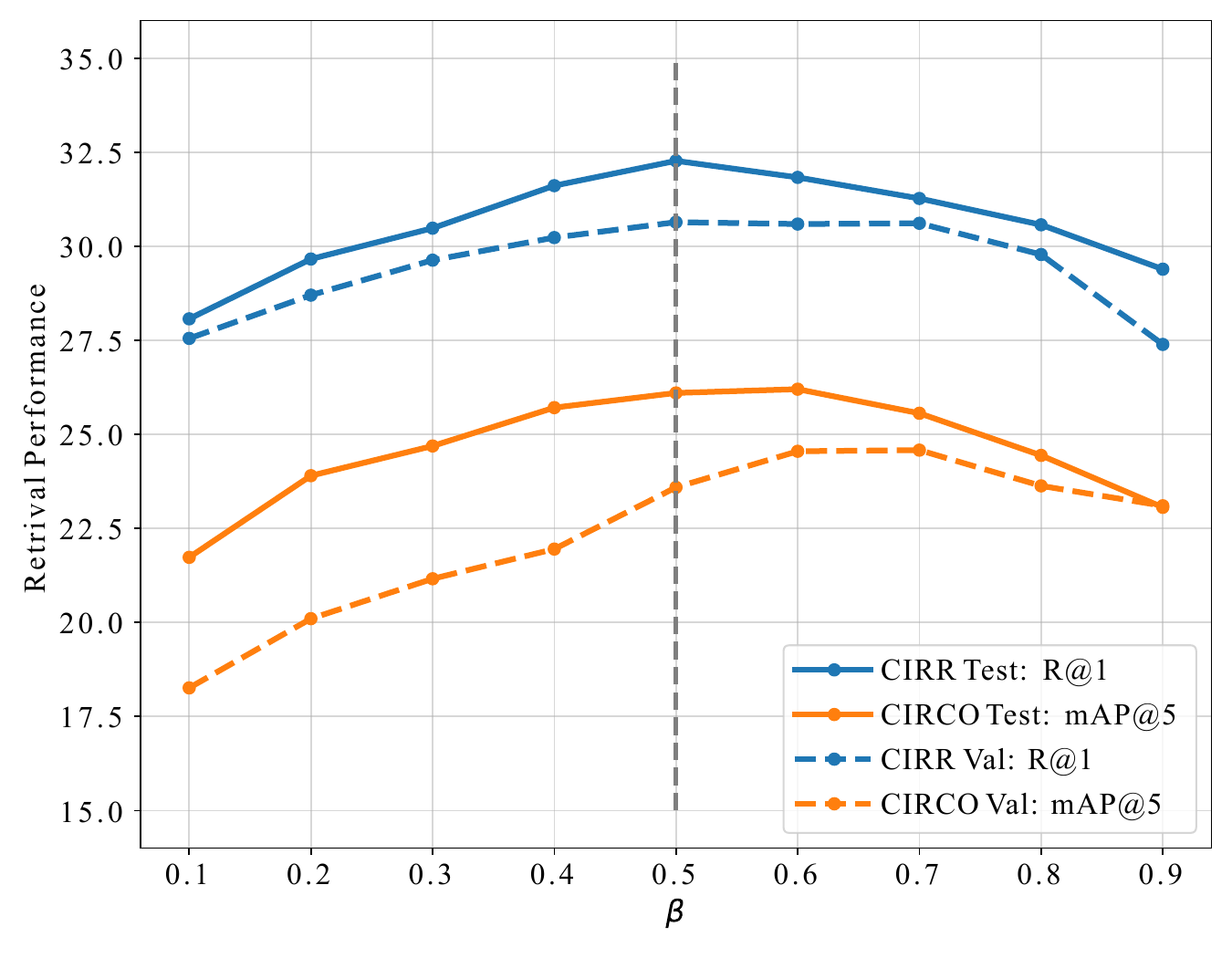}
\end{minipage}
\vspace{-2mm}
\hfill
\begin{minipage}[c]{0.48\textwidth}
\caption{\small
  {\bf Analysis of $\beta$.} 
  The parameter $\beta$ balances the contributions of real database images $V(I^i)$ and their synthetic counterparts $V(I^i_{\text{syn}})$ in Eq.~\ref{eq:features}.
  Based on studies on the CIRR and CIRCO validation sets using CLIP ViT-B/32, we set $\beta=0.5$ throughout our experiments, which not only yields competitive performance but also aligns with the principle of model simplicity by avoiding asymmetric weighting and reducing hyperparameter tuning.
  }
\label{fig:beta}
\end{minipage}
\vspace{-2mm}
\end{figure}

%
Furthermore, 
we test the weighting parameter $\beta$ in \cref{eq:features}, where $\beta$ and $(1-\beta)$ weight the features of real database images $V(\I^i)$ and their synthetic counterparts $V(\I_{syn}^i)$, respectively.
Based on the results of the CIRR and CIRCO validation sets with CLIP ViT-B/32 (\Cref{fig:beta}), we set $\beta=0.5$ throughout our experiments, as this balanced weighting not only achieves competitive performance but also simplifies the method by eliminating the need for hyperparameter tuning.

\begin{table}[t]
\centering
\small
\caption{\small
\textbf{Further analyses and ablation studies.}
We evaluate two potential enhancements to Paracosm: (1) replacing Qwen2.5-VL with GPT-4o for generating textual descriptions of mental images, and (2) incorporating brief textual descriptions of database images (w/ $\mathbf{t}^i_{brief}$) into the pipeline (see overview in \cref{fig:framework_suppl}).
However, neither change yields significant performance gains across benchmarks. 
These results indicate that Paracosm is already well-optimized, with limited room for improvement from using larger LMMs or augmenting database image descriptions.
We obtain these results using the CLIP ViT-B/32 VLM.
}
\vspace{-3mm}
\setlength{\tabcolsep}{5mm}
\scalebox{0.8}
{
    \begin{tabular}{lllcccccccccccccccc}
    \toprule

    \multicolumn{1}{c}{} & \multicolumn{2}{c}{CIRR} & \multicolumn{2}{c}{CIRCO} & \multicolumn{2}{c}{Fashion IQ} \\ 

    \cmidrule(lr){2-3}
    \cmidrule(lr){4-5}
    \cmidrule(lr){6-7}

    & \multicolumn{2}{c}{Recall@k} 
    & \multicolumn{2}{c}{mAP@k}  
    & \multicolumn{2}{c}{Average}  

    \\
   
     \textbf{Method} & k=1 & k=50 &  k=5 & k=50 & R@10 & R@50 \\

    \cmidrule(lr){1-1}
    \cmidrule(lr){2-3}
    \cmidrule(lr){4-5} 
    \cmidrule(lr){6-7}

    w/ GPT4o  & 31.86 & 92.41 & 26.44 & 30.85 & 26.65 & 46.14  \\
									
    w/ Qwen2.5-VL  & 32.27 & 92.60 & 26.10 & 30.45 & 26.14 & 46.39  \\

     w/ $\mathbf{t}^i_{brief}$ & 32.89 & 92.82 & 26.95 & 31.20 & 26.06 & 45.56 \\

\bottomrule

\end{tabular}
}
\label{tab:further_analysis_supply} 
\end{table}

\begin{table}[t]
\centering
\small
\caption{\small
\textbf{Robustness studies.}
We report Paracosm using different image generators (Qwen and LongCat) 
and different random seeds on the CIRR test set with the CLIP ViT-B/32 backbone.
The results, especially those in the {\setlength{\fboxsep}{1pt}\colorbox{col11}{pink rows}} for cross-generator configurations, confirm Paracosm's robustness to changes of image generators.
The bottom three rows demonstrate its robustness to random seeds adopted in these generative models. 
}
\vspace{-3mm}
\setlength{\tabcolsep}{4mm}
\scalebox{0.75}
{
    \begin{tabular}{llcccccc} 
        \toprule
         \textbf{Generator for query} & \textbf{Generator for database} &   R@1 &   R@5 &   R@10 &   R@50 \\
        \midrule
        LongCat (seed=43) & LongCat (seed=43) & 32.12 & 62.20 & 74.43 & 92.24\\
        \cellcolor{col11}LongCat (seed=43) & \cellcolor{col11}Qwen (seed=0) & \cellcolor{col11}33.01 & \cellcolor{col11}63.08 & \cellcolor{col11}75.25 & \cellcolor{col11}92.87 \\
        \cellcolor{col11}Qwen (seed=42) & \cellcolor{col11}LongCat (seed=43) & \cellcolor{col11}31.18 & \cellcolor{col11}61.49 & \cellcolor{col11}73.81 & \cellcolor{col11}91.71 \\
        Qwen (seed=42) & Qwen (seed=0) & 32.27 & 62.60 & 75.16  & 92.60 \\
        Qwen (seed=1) & Qwen (seed=1) & 32.02 & 61.78 & 74.07 & 91.78 \\
        Qwen (seed=2) & Qwen (seed=2) & 31.50 & 62.58 & 74.55 & 92.27 \\
        \bottomrule
    \end{tabular}
}
\label{tab:robussness_analysis_supply} 
\end{table}

\begin{table}[t]
\centering
\small
\caption{\small
\textbf{Detailed results corresponding to \Cref{fig:radar-chart}.} 
All results in this table are obtained using the OpenCLIP ViT-G/14 backbone to ensure fair comparison across methods.
We report quantitative results for methods on CIRR, CIRCO, and Fashion IQ, matching the visual summary in \Cref{fig:radar-chart}. For FashionIQ, we break down results by category: Shirt, Dress, and Toptee.
}
\vspace{-3mm}
\setlength{\tabcolsep}{0.4em}
\scalebox{0.735}
{
    \begin{tabular}{llccccccccccccc}
    \toprule
    \multicolumn{1}{l}{Dataset} &  \multicolumn{1}{l}{Subset} & \multicolumn{1}{l}{Metreic} 
    & \multicolumn{1}{c}{Pic2Word~\cite{Saito_2023_pic2word}} & \multicolumn{1}{c}{SEARLE~\cite{Baldrati_2023_searle}} & \multicolumn{1}{c}{LDRE~\cite{yang_2024_ldre}} & \multicolumn{1}{c}{CIReVL~\cite{Shyamgopal_2024_CIReVL}} & \multicolumn{1}{c}{OSrCIR~\cite{Tang_2025_OSrCIR}} & \multicolumn{1}{c}{Paracosm}  \\

    
    \cmidrule(lr){1-9}

    CIRR & -- & R@1  & 30.41 & 34.80 & 36.15 & 34.65 & 37.26 & \bf 39.30 \\

    CIRR & -- & R$_{\text{Subset}}$@1  & 69.92 & 68.72 & 68.82 & 67.95 & 69.22 & \bf 70.82 \\

    CIRCO & -- & mAP@5  & \ \ 5.54 & 13.20 & 31.12 & 26.77 & 25.62 & \bf 39.82 \\
    
    \multirow{3}{*}{Fashion IQ} & Shirt &  R@10  &  33.17  & 36.46 &  35.94 & 33.71 & 38.65 & \bf 40.48 \\

      & Dress & R@10 & 25.43 & 28.16  &  26.11 & 27.07 & 33.02 & \bf 33.17 \\

      & Toptee & R@10 & 35.24 & 39.83 &  35.42 & 35.80 & 41.04 & \bf 42.58 \\




    
\bottomrule

\end{tabular}
}
\vspace{-0mm}
\label{tab:radar_results_vitG_supply} 
\end{table}

\section{Open-Source Code}
\label{sec:open_source_code_suppl}
We provide demo code ({\tt demo.ipynb}) in our \href{https://github.com/leowangtong/Paracosm}{Github repository} for reproducibility.
The notebook implements the full Paracosm pipeline, 
including mental image generation, description generation, synthetic counterpart generation, and retrieval on the CIRCO validation set. 
We also provide {\tt README.md} with detailed guidelines on how to set up the environment and run the demo of Paracosm.



\section{Detailed Quantitative Results}
\label{sec: Detail_results_suppl}
This section provides additional results to complement the findings in the main paper.
\Cref{tab:radar_results_vitG_supply} presents the detailed numerical values corresponding to the radar chart in \Cref{fig:radar-chart}, enabling a quantitative comparison across methods and datasets.
Since AutoCIR \cite{cheng2025autocir} and CoTMR \cite{Sun_2025_cotmr} are implemented using ViT-B/32 and ViT-L/14 from OpenCLIP \cite{ilharco_gabriel_2021_openclip}, we provide a fair comparison under the same backbone family in \Cref{tab:openclip_results_suppl}.
Results show that Paracosm consistently outperforms both baselines across OpenCLIP ViT-B/32 and ViT-L/14, demonstrating the effectiveness of our approach independent of the specific VLM implementation.

\begin{table*}[t]
\centering
\small
\caption{\small
\textbf{Comparison with OpenCLIP-based methods under OpenCLIP ViT-B/32 and ViT-L/14 backbones.}
Since AutoCIR~\cite{cheng2025autocir} and CoTMR~\cite{Sun_2025_cotmr} are implemented with OpenCLIP~\cite{ilharco_gabriel_2021_openclip}, we provide a fair comparison under the same backbone family. Results are reported on CIRR (Recall@1/5/10), CIRCO (mAP@k), and Fashion-IQ (Recall@10) benchmarks. Best results are highlighted in \textbf{bold}, and second-best results are \underline{underlined}.
}
\vspace{-3mm}
\setlength{\tabcolsep}{2.mm}
\scalebox{0.65}
{
    \begin{tabular}{lllccccccccccccc}
    \toprule
    \multicolumn{3}{c}{} 
    & \multicolumn{2}{c}{CIRR} & \multicolumn{1}{c}{CIRCO} & \multicolumn{3}{c}{Fashion IQ}\\
    \cmidrule(lr){4-5}
    \cmidrule(lr){6-6}
    \cmidrule(lr){7-9}
    
    \textbf{Backbone} & \textbf{Method}  &  \textcolor{gray}{venue\&year}
    & \multicolumn{1}{c}{Recall@1} 
    & \multicolumn{1}{c}{Recall$_{\text{Subset}}$@1} 
    & \multicolumn{1}{c}{mAP@5}  
    & \multicolumn{1}{c}{Shirt R@10}  
    & \multicolumn{1}{c}{Dress R@10}  
    & \multicolumn{1}{c}{Toptee R@10}

    \\
 
    \cmidrule(lr){1-1}
    \cmidrule(lr){2-2}
    \cmidrule(lr){3-3} 
    \cmidrule(lr){4-4}
    \cmidrule(lr){5-5}
    \cmidrule(lr){6-6} 
    \cmidrule(lr){7-9}

    \multirow{3}{*}{ViT-B/32} & AutoCIR~\cite{cheng2025autocir}  & \textcolor{gray}{KDD'25} & 30.53 & 65.11 & 18.82 & 32.43 & 26.52 & 33.96 \\
									
    & CoTMR~\cite{Sun_2025_cotmr}  & \textcolor{gray}{ICCV'25} & \underline{31.50} & \underline{66.61} & \underline{22.23} & \underline{33.42} &  \textbf{31.09} & \underline{38.40} \\
    
    & \bf Paracosm  & \textcolor{gray}{ours} & \textbf{36.29} & \textbf{69.52} & \textbf{31.31} & \textbf{36.70} & \underline{29.40} & \textbf{39.32} \\

    \midrule 

    \multirow{3}{*}{ViT-L/14} & AutoCIR~\cite{cheng2025autocir}  & \textcolor{gray}{KDD'25} & 31.81 & 67.21 & 24.05 & 34.00 & 24.94 & 33.10 \\
									
    & CoTMR~\cite{Sun_2025_cotmr}  & \textcolor{gray}{ICCV'25} & \underline{35.02} & \underline{69.39} & \underline{27.61} & \underline{35.43} & \underline{31.18} & \underline{38.55} \\
    
    & \bf Paracosm  & \textcolor{gray}{ours} & \textbf{38.24} & \textbf{70.60} & \textbf{37.40} & \textbf{37.14} & \textbf{31.93} & \textbf{40.29} \\

\bottomrule

\end{tabular}
}
\vspace{-0mm}
\label{tab:openclip_results_suppl} 
\end{table*}

\begin{table*}[t]
\centering
\small
\caption{\small
\textbf{Ablation study.} All results in this table are obtained using the CLIP ViT-B/32 VLM \cite{radford2021clip}. To evaluate the contribution of each component, we conduct a comprehensive ablation study by removing one module at a time while keeping others unchanged. 
Per \cref{eq:features}, we focus on constructing features for {\setlength{\fboxsep}{1pt}\colorbox{col11}{multimodal queries}} and {\setlength{\fboxsep}{1pt}\colorbox{col33}{database images}}.
$(\I_{mental}, {\mathbf t}_{query}, {\mathbf t}_{mod})$ denotes (the generated mental image, description of the target image based on the query, modification text), while $(\I^i, \I_{syn}^i, {\mathbf t}^i_{detailed})$ denotes (database image, synthetic counterpart, detailed description of database image).
Paracosm's results are highlighted in \textbf{bold}.}
\setlength{\tabcolsep}{0.5em}
\scalebox{0.6}
{
    \begin{tabular}{lllccccccccccccccccc}
    \toprule
    \multicolumn{3}{c}{\cellcolor{col11} { multimodal query}} & \multicolumn{3}{c}{\cellcolor{col33} database images}
    & \multicolumn{7}{c}{CIRR} & \multicolumn{4}{c}{CIRCO}\\
    \cmidrule(lr){1-6}
    \cmidrule(lr){7-13}
    \cmidrule(lr){14-17}
    
    \cellcolor{col11}  & \cellcolor{col11}   & \cellcolor{col11}  & \cellcolor{col33} & \cellcolor{col33} &  \cellcolor{col33}
    & \multicolumn{4}{c}{Recall@k} 
    & \multicolumn{3}{c}{Recall$_{\text{Subset}}$@k} 
    & \multicolumn{4}{c}{mAP@k}  

    \\
   
     \cellcolor{col11}\textbf{$\I_{mental}$} & \cellcolor{col11}\textbf{${\mathbf t}_{query}$} & \cellcolor{col11}\textbf{${\mathbf t}_{mod}$} & \cellcolor{col33}\textbf{$\I^i$}  & \cellcolor{col33}\textbf{$\I_{syn}$} & \cellcolor{col33}\textbf{${\mathbf t}^i_{detailed}$} & k=1 &  k=5  & k=10 &  k=50  & k=1 &  k=2  & k=3 &  k=5 & k=10 &  k=25 & k=50 \\
    \cmidrule(lr){1-6}
    \cmidrule(lr){7-10}
    \cmidrule(lr){11-13}
    \cmidrule(lr){14-17}

     \cellcolor{col11}\checkmark  &  \cellcolor{col11} &  \cellcolor{col11} & \cellcolor{col33}\checkmark & \cellcolor{col33} & \cellcolor{col33} & 15.35 & 38.34 & 51.66 & 75.35 & 45.42 & 67.59 & 81.52 & \ \ 8.42 & \ \ 8.52 & \ \ 9.50 & 10.01 \\

     \cellcolor{col11}  &  \cellcolor{col11}\checkmark &  \cellcolor{col11} & \cellcolor{col33}\checkmark & \cellcolor{col33} & \cellcolor{col33} & 17.21 & 43.49 & 55.90 & 81.16 & 52.00 & 72.31 & 84.92 & 14.91 & 15.34 & 16.79 & 17.60 \\

     \cellcolor{col11}\checkmark &  \cellcolor{col11}\checkmark & \cellcolor{col11} & \cellcolor{col33}\checkmark & \cellcolor{col33} & \cellcolor{col33} & 18.80 & 44.96 & 58.84 & 82.65 & 50.39 & 72.80 & 83.93 & 13.71 & 13.89 & 15.19 & 15.92 \\

     \cellcolor{col11}\checkmark  &  \cellcolor{col11}\checkmark &  \cellcolor{col11}\checkmark & \cellcolor{col33}\checkmark & \cellcolor{col33} & \cellcolor{col33} & 27.93 & 57.11 & 70.29 & 90.31 & 61.88 & 80.70 & 90.70 & 18.29 & 18.69 & 20.45 & 21.27 \\

     \midrule

     \cellcolor{col11}\checkmark  &  \cellcolor{col11} &  \cellcolor{col11} & \cellcolor{col33} & \cellcolor{col33}\checkmark & \cellcolor{col33} & 17.64 & 41.93 & 54.46 & 77.37 & 48.36 & 69.11 & 83.21 & 10.06 & 10.43 & 11.44 & 12.03 \\

     \cellcolor{col11}  &  \cellcolor{col11}\checkmark &  \cellcolor{col11} & \cellcolor{col33} & \cellcolor{col33}\checkmark & \cellcolor{col33} & 17.21 & 43.93 & 56.29 & 80.10 & 50.94 & 71.81 & 84.53 & 13.58 & 13.92 & 15.44 & 16.30 \\

     \cellcolor{col11}\checkmark &  \cellcolor{col11}\checkmark & \cellcolor{col11} & \cellcolor{col33} & \cellcolor{col33}\checkmark & \cellcolor{col33} & 19.95 & 46.29 & 59.98 & 82.34 & 51.42 & 72.17 & 84.80 & 14.07 & 14.43 & 15.76 & 16.42 \\

     \cellcolor{col11}\checkmark  &  \cellcolor{col11}\checkmark &  \cellcolor{col11}\checkmark & \cellcolor{col33} & \cellcolor{col33}\checkmark & \cellcolor{col33} & 25.95 & 54.24 & 67.16 & 88.84 & 60.80 & 79.71 & 89.74 & 17.43 & 17.98 & 19.62 & 20.46 \\

     \midrule

     \cellcolor{col11}\checkmark  &  \cellcolor{col11} &  \cellcolor{col11} & \cellcolor{col33}\checkmark & \cellcolor{col33}\checkmark & \cellcolor{col33} & 20.82 & 48.48 & 60.68 & 82.72 & 51.33 & 72.24 & 84.94 & 13.48 & 13.79 & 15.24 & 15.89 \\

     \cellcolor{col11}  &  \cellcolor{col11}\checkmark &  \cellcolor{col11} & \cellcolor{col33}\checkmark & \cellcolor{col33}\checkmark & \cellcolor{col33} & 22.46 & 50.99 & 63.59 & 84.58 & 53.88 & 74.80 & 86.27 & 16.72 & 17.41 & 19.19 & 20.11 \\

     \cellcolor{col11}\checkmark &  \cellcolor{col11}\checkmark & \cellcolor{col11} & \cellcolor{col33}\checkmark & \cellcolor{col33}\checkmark & \cellcolor{col33} & 24.60 & 52.68 & 65.86 & 86.53 & 54.84 & 74.92 & 86.82 & 16.57 & 17.10 & 18.59 & 19.30 \\

     \cellcolor{col11}\checkmark  &  \cellcolor{col11}\checkmark &  \cellcolor{col11}\checkmark & \cellcolor{col33}\checkmark & \cellcolor{col33}\checkmark & \cellcolor{col33} & \textbf{32.27} & \textbf{62.60} & \textbf{75.16} & \textbf{92.60} & \textbf{65.16} & \textbf{83.25} & \textbf{92.34} & \textbf{26.10} & \textbf{27.02} & \textbf{29.29} & \textbf{30.45} \\

     \midrule

     \cellcolor{col11}\checkmark  &  \cellcolor{col11} &  \cellcolor{col11} & \cellcolor{col33}\checkmark & \cellcolor{col33}\checkmark & \cellcolor{col33}\checkmark & 22.31 & 50.94 & 63.86 & 85.01 & 52.96 & 73.71 & 85.98 & 13.88 & 14.48 & 15.87 & 16.57 \\

     \cellcolor{col11}  &  \cellcolor{col11}\checkmark &  \cellcolor{col11} & \cellcolor{col33}\checkmark & \cellcolor{col33}\checkmark & \cellcolor{col33}\checkmark & 22.19 & 50.02 & 62.36 & 84.43 & 53.37 & 75.64 & 86.60 & 13.66 & 14.32 & 15.81 & 16.57 \\

     \cellcolor{col11}\checkmark &  \cellcolor{col11}\checkmark & \cellcolor{col11} & \cellcolor{col33}\checkmark & \cellcolor{col33}\checkmark & \cellcolor{col33}\checkmark & 25.33 & 54.94 & 67.37 & 87.59 & 56.77 & 76.43 & 87.71 & 19.93 & 20.52 & 22.23 & 23.11 \\

     \cellcolor{col11}\checkmark  &  \cellcolor{col11}\checkmark &  \cellcolor{col11}\checkmark & \cellcolor{col33}\checkmark & \cellcolor{col33}\checkmark & \cellcolor{col33}\checkmark & 28.33 & 57.30 & 70.39 & 90.89 & 62.39 & 81.01 & 91.35 & 21.05 & 21.53 & 23.48 & 24.46 \\

\bottomrule

\end{tabular}
}
\label{tab:cirr_circo_results_ablation_suppl} 
\end{table*}

\end{document}